\documentclass[final]{cvpr}

\usepackage{times}
\usepackage{epsfig}
\usepackage{graphicx}
\usepackage{amsmath}
\usepackage{amssymb}
\usepackage{booktabs}
\usepackage{comment}
\usepackage{epigraph}

\usepackage{multirow}
\usepackage{hhline}
\usepackage{makecell}
\usepackage{bm}
\usepackage{xcolor}
\usepackage{subcaption}
\usepackage{soul}
\usepackage{array}
    \newcolumntype{P}[1]{>{\centering\arraybackslash}p{#1}}
    \newcolumntype{M}[1]{>{\centering\arraybackslash}m{#1}}
    \newcommand{\PreserveBackslash}[1]{\let\temp=\\#1\let\\=\temp}
    \newcolumntype{C}[1]{>{\PreserveBackslash\centering}p{#1}}
    \newcolumntype{R}[1]{>{\PreserveBackslash\raggedleft}p{#1}}
    \newcolumntype{L}[1]{>{\PreserveBackslash\raggedright}p{#1}}

\usepackage[pagebackref=true,breaklinks=true,colorlinks,bookmarks=false]{hyperref}

\makeatletter
\def\blfootnote{\gdef\@thefnmark{}\@footnotetext}
\makeatother

\def\cvprPaperID{2976} 
\def\confYear{CVPR 2021}

\begin{document}

\title{Learning by Aligning Videos in Time}

\author{Sanjay Haresh$^*$~~~~~Sateesh Kumar$^*$~~~~~Huseyin Coskun~~~~~Shahram N. Syed\\
Andrey Konin~~~~~M. Zeeshan Zia~~~~~Quoc-Huy Tran\\
\\
Retrocausal, Inc.\\
Seattle, WA\\
\url{www.retrocausal.ai}
}

\twocolumn[{%
\maketitle
\begin{center}
	\centering
		\includegraphics[width=0.95\linewidth, trim = 0mm 80mm 0mm 0mm, clip]{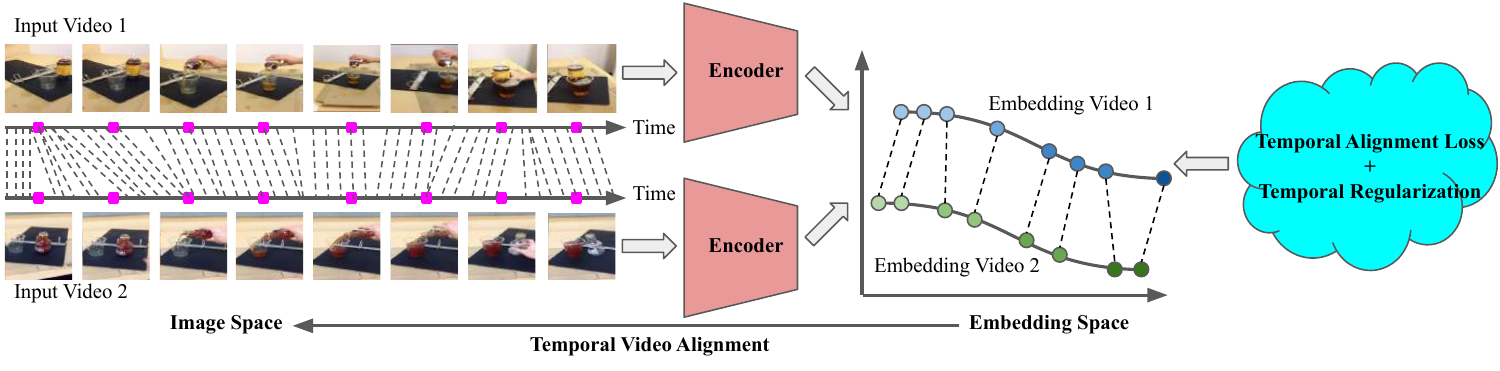}
    \vspace{-0.3cm}
    \captionof{figure}{We propose a self-supervised method to learn video representations by aligning videos in time, despite many differences between the videos such as appearance, motion, and viewpoint. We optimize the embedding space by using both the temporal alignment loss between the videos and the temporal regularization applied separately on each video. Our learned representations can be useful for many video-based temporal understanding tasks such as temporal video alignment.}
	\label{fig:teaser}
\end{center}%
}]

\begin{abstract}
We present a self-supervised approach for learning video representations using temporal video alignment as a pretext task, while exploiting both frame-level and video-level information. We leverage a novel combination of temporal alignment loss and temporal regularization terms, which can be used as supervision signals for training an encoder network. Specifically, the temporal alignment loss (i.e., \emph{Soft-DTW}) aims for the minimum cost for temporally aligning videos in the embedding space. However, optimizing solely for this term leads to trivial solutions, particularly, one where all frames get mapped to a small cluster in the embedding space. To overcome this problem, we propose a temporal regularization term (i.e., \emph{Contrastive-IDM}) which encourages different frames to be mapped to different points in the embedding space. Extensive evaluations on various tasks, including action phase classification, action phase progression, and fine-grained frame retrieval, on three datasets, namely \emph{Pouring}, \emph{Penn Action}, and \emph{IKEA ASM}, show superior performance of our approach over state-of-the-art methods for self-supervised representation learning from videos. In addition, our method provides significant performance gain where labeled data is lacking. Our code and labels are available on our research website: \url{https://retrocausal.ai/research/}.
\vspace{-0.2cm}
\end{abstract}

\section{Introduction}
\label{sec:introduction}

{\blfootnote{$^*$ indicates joint first author.\\ \{sanjay,sateesh,huseyin,shahram,andrey,zeeshan,huy\}@retrocausal.ai}}

\epigraph{There are just three problems in computer vision: registration, registration, and registration.}{\textit{Takeo Kanade}}
\vspace{-0.2cm}

Lukas-Kanade and Iterative Closest Point have been amongst the most ubiquitous building blocks in artificial perception literature. Yet spatio-temporal registration has received little attention in the present deep learning renaissance. Correspondingly, we add to a small number of recent approaches~\cite{dwibedi2019temporal, purushwalkam2020aligning} that have revived temporal alignment as a means of improving video representation learning. In order to learn perfect alignment of two videos, a learning algorithm must be able to disentangle phases of the activity in time while simultaneously associating visually similar frames in the two different videos. We demonstrate that learning in this manner generates representations that are effective for downstream tasks that rely on fine-grained temporal features.

In the context of using temporal alignment for learning video representations, some recent works~\cite{dwibedi2019temporal,purushwalkam2020aligning} use cycle-consistency losses to perform local alignment between individual frames. At the same time, some works have explored global alignment for  video classification and segmentation~\cite{chang2019d3tw,cao2020few}. We adapt such global alignment ideas for video representation learning in this work.

A few of approaches have been proposed for supervised action recognition~\cite{tran2015learning,carreira2017quo,wang2018non,tran2018closer} and action segmentation~\cite{farha2019ms,Li2020MSTCNMT}. Unfortunately, these approaches require fine-grained annotations which can be prohibitively expensive~\cite{richard2019temporal}. We note the seemingly infinite supply of public video data, and contrast it with the high cost of fine-grained annotation. This discrepancy emphasizes the importance of exploring self-supervised methods. We are further motivated by datasets and downstream tasks that specifically benefit from temporal alignment, such as video streams of semi-repetitive activities from manufacturing assembly lines to surgery rooms. It is desirable to measure the variability and anomalies~\cite{sultani2018real,haresh2020towards} across such datasets, where representations that optimize for temporal alignment may be highly performant. 

Our approach, \emph{L}earning by \emph{A}ligning \emph{V}ideos (\emph{LAV}), utilizes the task of temporally aligning videos for learning self-supervised video representations. Specifically, we use a differentiable version of an alignment metric which has been widely used in the time series literature, namely Dynamic Time Warping (DTW)~\cite{berndt1994using}. DTW is a global alignment metric, taking into account entire sequences while aligning. Unfortunately, in a self-supervised representation learning context, optimizing solely for DTW may converge to trivial solutions wherein the learned representations are not meaningful. To address this issue, we combine the above alignment metric with a regularization, as shown in Fig.~\ref{fig:teaser}. In particular, we propose a regularization term that optimizes for temporally disentangled representations, i.e., frames that are close in time are mapped to spatially nearby points in the embedding space and vice versa. 

In summary, our contributions include:
\vspace{-0.2cm}
\begin{itemize}
    \item We introduce a novel self-supervised method for learning video representations by temporally aligning videos as a whole, leveraging both frame-level and video-level cues.
    \vspace{-0.2cm}
    \item We adopt the classical DTW as our temporal alignment loss, while proposing a new temporal regularization. The two components have mutual benefits, i.e., the latter prevents trivial solutions, whereas the former leads to better performance.
    \vspace{-0.2cm}
    \item Our approach performs on par with or better than the state-of-the-art on various temporal understanding tasks on \emph{Pouring}, \emph{Penn Action}, and \emph{IKEA ASM} datasets. The best performance is sometimes achieved by combining our method with a recent work~\cite{dwibedi2019temporal}. Further, our approach offers significant accuracy gain when lacking labeled data.
    \vspace{-0.2cm}
    \item We manually annotate dense per-frame labels for 2123 videos of \emph{Penn Action}.
\end{itemize}

\section{Related Work}
\label{sec:relatedwork}

In this section, we review recent literature in self-supervised learning with a focus on image and video data.

\noindent \textbf{Image-Based Self-Supervised Representation Learning.} Early self-supervised representation learning methods explore image content as supervision signals. They propose pretext tasks based on artificial image cues as labels and train deep networks for solving those tasks~\cite{larsson2016learning,larsson2017colorization,noroozi2017representation,liu2018leveraging,kim2018learning,gidaris2018unsupervised,carlucci2019domain,feng2019self}. These pretext tasks include objectives such as  image colorization~\cite{larsson2016learning,larsson2017colorization}, object counting~\cite{noroozi2017representation,liu2018leveraging}, solving jigsaw puzzles~\cite{kim2018learning,carlucci2019domain}, and predicting image rotations~\cite{gidaris2018unsupervised,feng2019self}. Even earlier approaches learn representations simply by reconstructing the input image~\cite{hinton1994autoencoders} or recovering it from noise~\cite{vincent2008extracting}. In this work, we focus on self-supervised representation learning from videos, which leverages both spatial and temporal information in videos.

\noindent \textbf{Video-Based Self-Supervised Representation Learning.} With the advent of deep architectures for video understanding~\cite{tran2015learning,carreira2017quo,wang2018non,tran2018closer}, various pretext tasks have been introduced as supervision signals for self-supervised representation learning from videos. One popular class of methods learn representations by predicting future frames~\cite{srivastava2015unsupervised,vondrick2016generating,ahsan2018discrimnet,diba2019dynamonet} or forecasting their encoding features~\cite{han2019video,kim2019self,gammulle2019predicting}. Another group of methods leverage temporal information, for example, temporal order and temporal coherence are used as labels in~\cite{misra2016shuffle,lee2017unsupervised,fernando2017self, xu2019self,choi2020shuffle} and~\cite{hadsell2006dimensionality,mobahi2009deep,bengio2009slow,zou2011unsupervised,zou2012deep,goroshin2015unsupervised} respectively. Recently, Donglai et al.~\cite{wei2018learning} train a deep model for classifying temporal direction, while Sermanet et al.~\cite{sermanet2018time} learn representations via consistency across different viewpoints  and neighboring frames. The above methods usually optimize over a single video at a time, whereas our approach jointly optimizes over a pair of videos at once, potentially extracting more information from both videos.

\noindent \textbf{Temporal Video Alignment.} There exists a lot of literature on time series alignment, yet only a few ideas have been carried over to aligning videos. Unfortunately, traditional methods for time series alignment, for example, DTW~\cite{berndt1994using}, are not differentiable and hence can not be directly used for training neural networks. To address this weakness, a smooth approximation of DTW, namely Soft-DTW, is introduced in~\cite{cuturi2017soft}. More recently, Soft-DTW formulations have been used in a weakly supervised setting for aligning a video to a transcript~\cite{chang2019d3tw} or in a few-shot supervised setting for aligning videos~\cite{cao2020few}. In the present paper, we adapt Soft-DTW for learning self-supervised representations from videos, using temporal video alignment as the pretext task. The closest work to ours is Temporal Cycle Consistency (TCC)~\cite{dwibedi2019temporal}, which learns self-supervised representations by finding frame correspondences across videos. While TCC aligns each frame separately, our approach aligns the video as a whole, leveraging both frame-level and video-level cues. 
\section{Our Approach}
\label{sec:method}

\begin{figure*}[t]
	\centering
		\includegraphics[width=0.95\linewidth, trim = 0mm 55mm 0mm 0mm, clip]{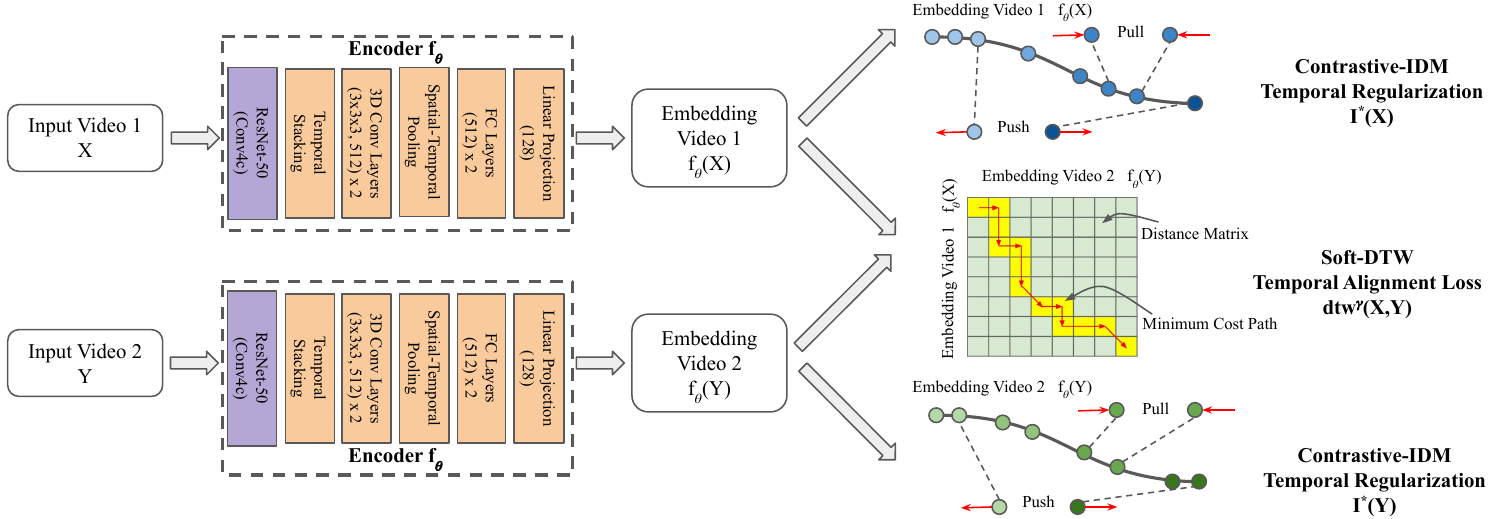}
	\vspace{-0.3cm}
	\caption{Given input videos $X$ and $Y$, we feed them to the encoder $f_\theta$ to obtain the embedding videos $f_\theta(X)$ and $f_\theta(Y)$. We optimize the encoder parameters $\theta$ by applying the Soft-DTW temporal alignment loss $dtw^\gamma(X,Y)$ and the Contrastive-IDM temporal regularization $I^*(X)$ and $I^*(Y)$ on the embedding videos $f_\theta(X)$ and $f_\theta(Y)$.}
	\label{fig:method}
	\vspace{-0.2cm}
\end{figure*}

In this section, we discuss our main contribution which is a self-supervised method to learn video representations via temporal video alignment. Specifically, we learn an embedding space where two videos with similar contents can be conveniently aligned in time. We first aim to optimize the embedding space solely for the global alignment cost between the two videos, which can lead to trivial solutions. To overcome this problem, we regularize the embedding space such that for each input video, temporally close frames are mapped to nearby points in the embedding space, whereas temporally distant frames are correspondingly mapped far away in the embedding space. Fig.~\ref{fig:method} shows an overview of our loss and regularization (right) and our encoder (left). Below we first define some notations and then provide the details of our temporal alignment loss, temporal regularization, final loss, and encoder network in Secs.~\ref{sec:loss},~\ref{sec:regularization},~\ref{sec:final}, and~\ref{sec:encoder} respectively. 

\noindent \textbf{Notations.} We denote the embedding function as $f_\theta$, namely a neural network with parameters $\theta$. Our method takes as input two videos $X = \{x_1, x_2, \dots, x_n\}$ and $Y = \{y_1, y_2, \dots, y_m\}$, where $n$ and $m$ are the numbers of frames in $X$ and $Y$ respectively. For a frame $x_i$ in $X$ and $y_j$ in $Y$, the embedding frames of $x_i$ and $y_j$ are written as $f_\theta(x_i)$ and $f_\theta(y_j)$ respectively. In addition, we denote $f_\theta(X) = \{f_\theta(x_1), f_\theta(x_2), \dots, f_\theta(x_n)\}$ and $f_\theta(Y) = \{f_\theta(y_1), f_\theta(y_2), \dots, f_\theta(y_m)\}$ as the embedding videos of $X$ and $Y$ respectively. 

\subsection{Temporal Alignment Loss}
\label{sec:loss}

We adopt the classical DTW discrepancy~\cite{berndt1994using} as our temporal alignment loss. DTW has been widely used with non-visual data, such as time series, and has just recently been applied to video data, but in a weakly supervised setup for video-to-transcript alignment~\cite{chang2019d3tw} or in a few-shot supervised setup for video alignment~\cite{cao2020few}. Unlike~\cite{chang2019d3tw,cao2020few}, we explore the use of DTW for self-supervised video representation learning by leveraging temporal video alignment as the pretext task.

Given two input videos $X$ and $Y$ and their embedding videos $f_\theta(X)$ and $f_\theta(Y)$, we can compute the distance matrix $D \in \mathbb{R}^{n \times m}$ with each entry written as $D(i, j) = ||f_\theta(x_i) - f_\theta(y_j)||^2$. DTW calculates the alignment cost between $X$ and $Y$ by finding the minimum cost path in $D$:
\begin{equation}
\label{eq:dtw}
    dtw(X, Y) = min_{A \in A_{n,m}} \langle A, D \rangle.
\end{equation}
Here, $A_{n,m} \subset \{0,1\}^{n \times m}$ is the set of all possible (binary) alignment matrices, which correspond to  paths from the top-left corner of $D$ to the bottom-right corner of $D$ using only $\{\downarrow, \rightarrow, \searrow\}$ moves. $A \in A_{n,m}$ is a typical alignment matrix, with $A(i,j) = 1$ if $x_i$ in $X$ is aligned with $y_j$ in $Y$. DTW can be computed using dynamic programming, particularly solving the below cumulative distance function:
\begin{equation}
\label{eq:dp}
    r(i,j) = D(i,j) + min\{r(i-1,j),r(i,j-1),r(i-1,j-1)\}.
\end{equation}

Due to the non-differentiable $min$ operator, DTW is not differentiable and unstable when used in an optimization framework. We therefore employ a continuous relaxation version of DTW, namely \emph{Soft-DTW}, proposed by~\cite{cuturi2017soft}. In particular, Soft-DTW replaces the discrete $min$ operator in DTW by the smoothed $min^\gamma$ one, defined as:
\begin{equation}
\label{eq:softmin}
    min^{\gamma} \{a_1, a_2, ..., a_n\} = -\gamma \log \sum_{i=1}^n e^{\frac{-a_i}{\gamma}},
\end{equation}
where $\gamma > 0$ is a smoothing parameter. Soft-DTW returns the alignment cost between $X$ and $Y$ by finding the soft-minimum cost path in $D$, which can be written as:
\begin{equation}
\label{eq:softdtw}
    dtw^\gamma(X,Y) = min^{\gamma}_{A \in A_{n,m}} \langle A, D \rangle.
\end{equation}
Note that since the smoothed $min^\gamma$ operator converges to the discrete $min$ one when $\gamma$ approaches 0, Soft-DTW produces similar results as DTW when $\gamma$ is near 0. In addition, although using $min^\gamma$ does not make the objective convex, it does help the optimization by enabling smooth gradients and providing better optimization landscapes.

\subsection{Temporal Regularization}
\label{sec:regularization}

Since (Soft-)DTW measures the (soft-)minimum cost path in $D$, optimizing for (Soft-)DTW alone can result in trivial solutions, wherein all the entries in $D$ are close to 0, as we will show later in Sec.~\ref{sec:ablation}. In other words, all the frames in $X$ and $Y$ are mapped to a small cluster in the embedding space. To avoid that, we opt to add a temporal regularization, which is applied separately on $f_\theta(X)$ and $f_\theta(Y)$. Below we discuss our regularization for $f_\theta(X)$ only, while the same one can be applied for $f_\theta(Y)$.

Motivated by~\cite{su2017order}, we adapt Inverse Difference Moment (IDM)~\cite{conners1980theoretical} as our regularization, which can be written as:
\vspace{-0.2cm}
\begin{align}
\label{eq:idm_max}
    I(X) = \sum_{i=1}^{n} \sum_{j=1}^{n} W(i,j) S_X(i,j)\\
    W(i,j) =\frac{1}{(i - j)^2+1} \nonumber,
\end{align}
where $S_X \in \mathbb{R}^{n \times n}$ is the self-similarity matrix of $f_\theta(X)$. Maximizing Eq.~\ref{eq:idm_max} encourages temporally close frames in $X$ (with large $W(i,j)$) to be mapped to nearby points in the embedding space (with large $S_X(i,j)$). Unlike~\cite{su2017order}, which applies IDM on the transport matrix between two (skeleton) sequences, we apply IDM separately on each (video) sequence. To be used as a loss function, we convert the above maximization objective to the below minimization:
\vspace{-0.2cm}
\begin{align}
\label{eq:idm_min}
    \bar{I}(X) & = \sum_{i=1}^{n} \sum_{j=1}^{n} \bar{W}(i,j)(-D_X(i,j))\\
    & \bar{W}(i,j) =(i - j)^2+1, \nonumber
\end{align}
where $D_X \in \mathbb{R}^{n \times n}$ is the self-distance matrix of $f_\theta(X)$, and is defined as $D_X(i, j) = ||f_\theta(x_i) - f_\theta(x_j)||^2$. Minimizing Eq.~\ref{eq:idm_min} encourages temporally close frames in $X$ (with small $\bar{W}(i,j)$) to be mapped to nearby points in the embedding space (with small $D_X(i,j)$).

However, we notice one problem with the above IDM regularization, in particular, it treats temporally close and far way frames in similar ways. In Eq.~\ref{eq:idm_max}, it maximizes similarities between temporally far away frames, though with smaller weights. Similarly, for Eq.~\ref{eq:idm_min}, it still maximizes distances between temporally close frames, though with smaller weights. To address that, we propose separate terms for temporally close and far away frames. Specifically, we introduce a contrastive version of Eq.~\ref{eq:idm_min}, which we call \emph{Contrastive-IDM}, as our regularization:
\vspace{-0.2cm}
\begin{equation}
\label{eq:idm_contrastive}
\begin{split}
    I^*(X) = \sum_{i=1}^{n} \sum_{j=1}^{n} y_{ij} \bar{W}(i,j)  max(0,\lambda - D_X(i,j)) \\
    + (1 - y_{ij}) W(i,j) D_X(i,j),\\
    y_{ij} = \begin{cases}
        1 ,& |i - j| > \sigma\\
        0, & |i - j| \leq \sigma\\
    \end{cases}.\\
\end{split}
\end{equation}
Here, $\sigma$ is a window size for separating temporally far away frames ($y_{ij}$ = 1 or \emph{negative} pairs) and temporally close frames ($y_{ij}$ = 0 or \emph{positive} pairs) and $\lambda$ is a margin parameter. Contrastive-IDM encourages temporally close frames (positive pairs) to be nearby in the embedding space, while penalizing temporally far away frames (negative pairs) when the distance between them is smaller than margin $\lambda$ in the embedding space. Note that, if we drop the weights $\bar{W}(i,j)$ and $W(i,j)$ in Eq.~\ref{eq:idm_contrastive}, it becomes equivalent to Slow Feature Analysis (SFA), also referred to as temporal coherence~\cite{hadsell2006dimensionality,mobahi2009deep,goroshin2015unsupervised}, which treats all pairs equally. We would emphasize that, leveraging temporal information by adding weights to different pairs based on their temporal gaps leads to performance gain, as we will show in Sec.~\ref{sec:ablation}.

\subsection{Final Loss}
\label{sec:final}

Our final loss is a combination of Soft-DTW alignment loss in Eq.~\ref{eq:softdtw} and Contrastive-IDM regularization in Eq.~\ref{eq:idm_contrastive}:
\begin{equation}
\label{eq:finalloss}
    L(X, Y) = dtw^\gamma(X, Y) + \alpha (I^*(X) + I^*(Y)).
\end{equation}
Here, $\alpha$ is the weight for the regularization. The final loss encourages embedding videos to have minimum alignment costs while encouraging discrepancies among embedding frames. Both the alignment loss and the regularization are differentiable and can be optimized using backpropagation.

\subsection{Encoder Network}
\label{sec:encoder}

We use ResNet-50~\cite{he2016deep} as our backbone network and extract features from the output of the $Conv4c$ layer. The extracted features have dimensions of $14 \times 14 \times 1024$. We then stack $k$ context frame features along the temporal dimension for each frame. Next, the combined features are passed through two 3D convolutional layers for aggregating temporal information. It is then followed by a 3D global max pooling layer, two fully-connected layers, and a linear projection layer to output embedding frames, with each having 128 dimensions. We resize input video frames to $224 \times 224$ before feeding to our encoder network.
 \section{Datasets, Annotations, and Metrics}
\label{sec:datasetsmetrics}

\begin{figure}[t]
	\centering
		\includegraphics[width=1.0\linewidth, trim = 0mm 120mm 145mm 0mm, clip]{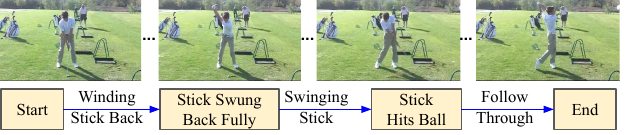}
    \vspace{-0.3cm}
	\caption{We annotate dense frame-wise labels (i.e., key events and phases) for 2123 videos of \emph{Penn Action}. In the above \emph{Golf Swing} video, key events represent specific events, e.g., \emph{Stick Swung Back Fully}, while phases are periods between key events, e.g., \emph{Winding Stick Back}.}
	\label{fig:annotation}
	\vspace{-0.2cm}
\end{figure}

\noindent \textbf{Datasets and Annotations.} We use three datasets, namely \emph{Pouring}~\cite{sermanet2018time}, \emph{Penn Action}~\cite{zhang2013actemes}, and \emph{IKEA ASM}~\cite{ben2020ikea}. While \emph{Pouring} videos capture human hands interacting with objects, \emph{Penn Action} and \emph{IKEA ASM} videos show humans playing sports and assembling furniture respectively. We manually annotate dense frame-wise labels (i.e., key events and phases) for \emph{Penn Action} using the same protocol of~\cite{dwibedi2019temporal}, since the authors of~\cite{dwibedi2019temporal} do not release them. See Fig.~\ref{fig:annotation} for an example. For \emph{Pouring} and \emph{IKEA ASM}, we obtain the labels from the authors of~\cite{dwibedi2019temporal} and~\cite{ben2020ikea} respectively. Actions/videos in \emph{IKEA ASM} (17 phases) are more complicated/longer than those in \emph{Pouring} (5 phases) and \emph{Penn Action} (2-6 phases). We use the training/validation splits from the original datasets. For \emph{Pouring}, we use all videos (70 for training, 14 for validation). Following~\cite{dwibedi2019temporal}, we use 13 actions of \emph{Penn Action} (for each action, 40-134 videos for training, 42-116 videos for validation). For \emph{IKEA ASM}, we use all \emph{Kallax\_Drawer\_Shelf} videos (61 for training, 29 for validation).

\noindent \textbf{Evaluation Metrics.} We use four evaluation metrics computed on the validation set. The network is first trained on the training set and then frozen. Next, an SVM classifier or linear regressor is trained on top of the frozen network features (without any fine-tuning of the network). For all metrics, a high score means a better model. We summarize the metrics below:
\vspace{-0.2cm}
\begin{itemize}
    \item \emph{Phase Classification:} is the average per-frame phase classification accuracy, implemented by training an SVM classifier on top of the frozen network features to predict the phase labels.
    \vspace{-0.2cm}
    \item \emph{Phase Progression~\cite{draper1998applied,wiki:Coefficient_of_determination}:} measures the prowess of representations learnt to predict action progress temporally, implemented by training a linear regressor on top of the frozen network features to predict the phase progression values (defined using the key event labels).
    \vspace{-0.2cm}
    \item \emph{Kendall's Tau~\cite{kendall1938new,wiki:Kendall_rank_correlation_coefficient}:} measures how well videos are aligned temporally if we use nearest neighbor matching. It does not require any labels for evaluation.
    \vspace{-0.2cm}
    \item \emph{Average Precision:} is the fine-grained frame retrieval accuracy, computed as the ratio of the retrieved frames with the same phase labels as the query frame.
\end{itemize}
We follow~\cite{dwibedi2019temporal} to use the first three metrics above, while we add the last metric for our fine-grained frame retrieval experiments in Sec.~\ref{sec:retrival}. Phase Progression and Kendall's Tau assume no repetitive frames/labels in a video.

\section{Experiments}
\label{sec:experiments}

In this section, we benchmark our approach (namely \emph{LAV}, short for \emph{L}earning by \emph{A}ligning \emph{V}ideos) against state-of-the-art methods for video-based self-supervised representation learning on various temporal understanding tasks on \emph{Pouring}, \emph{Penn Action}, and \emph{IKEA ASM} datasets. 

\noindent \textbf{Implementation Details.} We use the same encoder in Sec.~\ref{sec:encoder} for all methods for \emph{Pouring} and \emph{Penn Action} experiments. For \emph{IKEA ASM} experiments, since the actions are more complex, we opt to extract features from the output of the $Conv5c$ layer (instead of $Conv4c$) for all methods. We initialize ResNet-50 layers with pre-trained weights for ImageNet classification, while remaining layers are initialized randomly. We L2-normalize the frame-embeddings before feeding them to our loss (LAV). We use ADAM optimization~\cite{kingma2014adam} with a learning rate of $10^{-4}$ and a weight decay of $10^{-5}$.  We minimize our final loss in Eq.~\ref{eq:finalloss} computed over all video pairs in the training set. We randomly pair videos of the same action, \emph{regardless} of their viewpoints. For datasets with a single action (e.g., \emph{Pouring} and \emph{IKEA ASM}), videos are randomly paired. For datasets with many actions (e.g., \emph{Penn Action}), videos of the same action are randomly paired. For each video pair, we calculate the final loss using $p$ sampled frames from each video, i.e., we divide a video into $p$ uniform chunks and randomly sample one frame per chunk. We implement our network and loss in PyTorch~\cite{paszke2017automatic}. For more details, please refer to supplementary materials.

\noindent \textbf{Competing Methods.} Below are the competing methods:
\vspace{-0.2cm}
\begin{itemize}
    \item \emph{Self-Supervised Learning:} We compare LAV with recent self-supervised video representation learning methods, namely SAL~\cite{misra2016shuffle}, TCN~\cite{sermanet2018time}, and~TCC\cite{dwibedi2019temporal}.
    \vspace{-0.2cm}
    \item \emph{Fully-Supervised Learning:} We test LAV against a fully-supervised method with explicit supervision. Specifically, following~\cite{dwibedi2019temporal}, we train a network on the downstream task by attaching a 1-layer classifier to the encoder in  Sec.~\ref{sec:encoder}.
    \vspace{-0.2cm}
    \item \emph{Random/ImageNet Features:} For completeness, we include the results obtained by using random features or pre-trained features for ImageNet classification.
\end{itemize}

\subsection{Ablation Study Results}
\label{sec:ablation}

\begin{table}
\vspace{-0.15cm}
\begin{minipage}[b]{1.0\linewidth}
\centering

\begin{tabular}{c|c|c|c|c}
\specialrule{1pt}{1pt}{1pt}

& \small{\textbf{Loss}} & \small{\textbf{Classification}} &  \small{\textbf{Progress}} & \bm{$\tau$} \\
\midrule
\multirow{4}{*}{\rotatebox[origin=c]{90}{\scriptsize{\textbf{Individual}}}}

& \small{S-DTW}~\cite{cuturi2017soft} & 48.35 & 0.2770 & 0.2144 \\

 & \small{IDM} (Eq.~\ref{eq:idm_min}) & 48.16 & 0.7241 & 0.5835 \\

& \small{SFA}~\cite{hadsell2006dimensionality} & 92.20 & \underline{0.7533} & 0.8093  \\
& \small{C-IDM} (Eq.~\ref{eq:idm_contrastive}) & \underline{92.82} & 0.7477 & \underline{0.8318} \\

\midrule
\multirow{3}{*}{\rotatebox[origin=c]{90}{\scriptsize{\textbf{Combined}}}}
 & \small{S-DTW + IDM}  & 
 68.73 & 0.6551 & 0.6408 \\

& \small{S-DTW + SFA}  & 91.63 & 0.7146 & 0.8069 \\

& \small{S-DTW + C-IDM}  & \textbf{92.84} & \textbf{0.8054} & \textbf{0.8561}  \\ 

\specialrule{1pt}{1pt}{1pt}

\end{tabular}

\caption{Ablation studies of individual losses (top) and combined losses (bottom). S-DTW and C-IDM denote Soft-DTW and Contrastive-IDM respectively. Best results are in \textbf{bold}, while second best ones are \underline{underlined}.}
\label{tab:results_ind_abl}
\vspace{-0.15cm}
\end{minipage}

\end{table}

\begin{figure}[t]
	\centering
		\includegraphics[width=0.95\linewidth, trim = 0mm 90mm 120mm 0mm, clip]{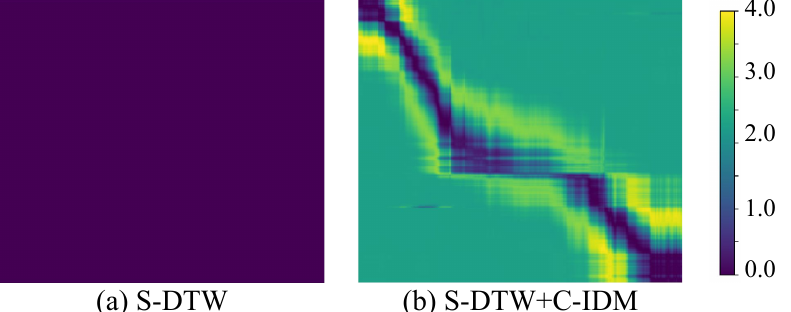}
    \vspace{-0.3cm}
	\caption{Distance matrices between embedding frames of two \emph{Pouring} videos learned by various losses. S-DTW denotes Soft-DTW, while C-IDM means Contrastive-IDM.}
	\label{fig:distmat}
	\vspace{-0.2cm}
\end{figure}

Here, we perform ablation studies on \emph{Pouring} dataset to show the effectiveness of our design choices in Sec.~\ref{sec:method}.

\noindent \textbf{Performance of Individual Losses.} We first study the performance of individual components of our approach, i.e., Soft-DTW and Contrastive-IDM, as separate baselines. Also, we include other methods such as IDM in Eq.~\ref{eq:idm_min} and an SFA approach proposed in~\cite{hadsell2006dimensionality}. Tab.~\ref{tab:results_ind_abl} (top) presents the quantitative results. We observe that the model trained with Soft-DTW alone achieves the lowest accuracy across all metrics. In fact, it has similar classification accuracy to that of random features in Tab.~\ref{tab:results_clf_sep} (i.e., $48.35\%$ vs. $45.10\%$), which shows that training solely with Soft-DTW yields trivial solutions and the network is unable to learn any useful representations. This is also confirmed by plotting the distance matrix between the embedding frames learned solely with Soft-DTW in Fig.~\ref{fig:distmat}(a), where all entries are near zero. In other words, the frames are mapped to a small cluster in the embedding space. Moreover, it can be seen from Tab.~\ref{tab:results_ind_abl} (top) that Contrastive-IDM outperforms IDM by significant margins on all metrics (e.g., for Kendall's Tau, $0.8318$ vs. $0.5835$), showing the advantage of using separate terms for temporally close and far away frame pairs. Lastly, although SFA and Contrastive-IDM have competitive performances on classification and progression, Contrastive-IDM outperforms SFA significantly on Kendall's Tau (i.e., $0.8318$ vs. $0.8093$), supporting our idea of adding weights to different frame pairs based on their temporal gaps.

\noindent \textbf{Performance of Combined Losses.} We now study the impact of adding IDM, SFA, or Contrastive-IDM as regularization to Soft-DTW. Tab.~\ref{tab:results_ind_abl} (bottom) presents the quantitative results. From the results, the addition of regularization boosts the performance of Soft-DTW significantly across all metrics (e.g., for progression, $0.2770$ for Soft-DTW vs. $0.8054$ for Soft-DTW+Contrastive-IDM). More importantly, utilizing our proposed Contrastive-IDM as regularization leads to the best performance across all metrics, outperforming using IDM or SFA as regularization by significant margins, especially on progression and Kendall's Tau (e.g., for progression, $0.8054$ for Soft-DTW+Contrastive-IDM vs. $0.6551$ and $0.7146$ for Soft-DTW+IDM and Soft-DTW+SFA respectively). This validates our ideas of separating temporally close and far away frame pairs, as well as leveraging temporal gaps to weight the frame pairs accordingly. We also visualize the distance matrix between the embedding frames learned with Soft-DTW+Contrastive-IDM in Fig.~\ref{fig:distmat}(b), where entries have diverse values. Below, we use Soft-DTW+Contrastive-IDM as our method (LAV).

\subsection{Phase Classification Results}
\label{sec:classification}

\begin{table}
\vspace{-0.15cm}
\begin{minipage}[b]{1.0\linewidth}
\centering

{%

\begin{tabular}{l|l|ccc}

\specialrule{1pt}{1pt}{1pt}

  & \textbf{\% of labels $\rightarrow$} & \textbf{0.1} & \textbf{0.5} & \textbf{1.0} \\
%
\midrule
\multirow{7}{*}{\rotatebox[origin=c]{90}{\thead{\textbf{Pouring}}}}
& \small{Supervised Learning}
 
&  72.44 & 89.57  &  92.86 \\

& \small{Random Features}
 
&   43.84 & 44.52 & 45.10
\\

& \small{Imagenet Features}
 
&   52.40 & 71.10 & 78.46
 \\
\cline{2-5}
& SAL~\cite{misra2016shuffle}
 
&  87.63 & 87.58 & 88.81\\

& TCN~\cite{sermanet2018time}
&  89.67 & 87.32 & 89.53 \\

& TCC~\cite{dwibedi2019temporal} 
 
&  90.65 & 91.11 & 91.53 \\

& LAV (Ours) 
&  \underline{91.61} & \textbf{92.82} & \underline{92.84} \\

& LAV + TCC (Ours)
 
&  \textbf{92.78} & \underline{92.56} & \textbf{93.07}\\


\midrule
\multirow{7}{*}{\rotatebox[origin=c]{90}{\thead{\textbf{Penn} \textbf{Action}}}}

& \small{Supervised Learning}
 
&   68.92 & 81.17 & 84.34
 \\
 
& \small{Random Features}
 
&   47.05 & 47.19 & 47.65 
\\

& \small{Imagenet Features}
 
&   46.66  & 56.39 & 60.65

 \\
\cline{2-5}
& SAL~\cite{misra2016shuffle}
 
&  79.94 & 81.11 & 81.79
  \\

& TCN~\cite{sermanet2018time}
 
&  81.99 & 82.64 & 82.78
   \\

& TCC~\cite{dwibedi2019temporal}
 
&  79.72 & 81.12 & 81.35
  \\

& LAV(Ours)
 
&  \textbf{83.56} & \textbf{83.95} & \textbf{84.25}
  \\
  
& LAV + TCC (Ours)
 
&  \underline{83.21} & \underline{83.79} & \underline{84.12} \\

\midrule
\multirow{7}{*}{\rotatebox[origin=c]{90}{\textbf{\thead{IKEA ASM}}}}

& \small{Supervised Learning}
 
& 21.76 & 30.26 & 33.81
 \\
 
& \small{Random Features}
 
&  17.89 & 17.89 & 17.89
\\

& \small{Imagenet Features}
 
&    18.05 & 19.27 & 19.50
\\
\cline{2-5}
& SAL~\cite{misra2016shuffle}
&  21.68 & 21.72 & 22.14
\\

& TCN~\cite{sermanet2018time}
&   \underline{25.17} & 25.70 & 26.80
\\

& TCC~\cite{dwibedi2019temporal}
&     24.74 & 25.22 & 26.46
\\

& LAV (Ours)
&  \textbf{29.78} & \underline{29.85} & \underline{30.43}
   \\

& LAV + TCC (Ours)
&  24.58 & \textbf{30.47} & \textbf{30.51}
   \\

\specialrule{1pt}{1pt}{1pt}
\end{tabular}

}

\caption{Phase classification results. Best results are in \textbf{bold}, while second best ones are \underline{underlined}.}
\label{tab:results_clf_sep}
\vspace{-0.15cm}
\end{minipage}

\end{table}

\begin{figure}
\begin{minipage}[b]{1.0\linewidth}
    \centering
    \begin{subfigure}[t]{0.5\textwidth}
        \centering
        \includegraphics[height=1.25in]{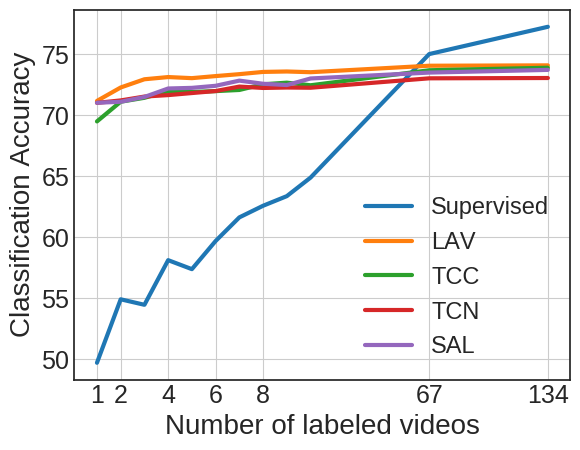}
        \caption{Bowling}
        \label{fig:fs_bowl}
    \end{subfigure}%
    \begin{subfigure}[t]{0.5\textwidth}
        \centering
        \includegraphics[height=1.25in]{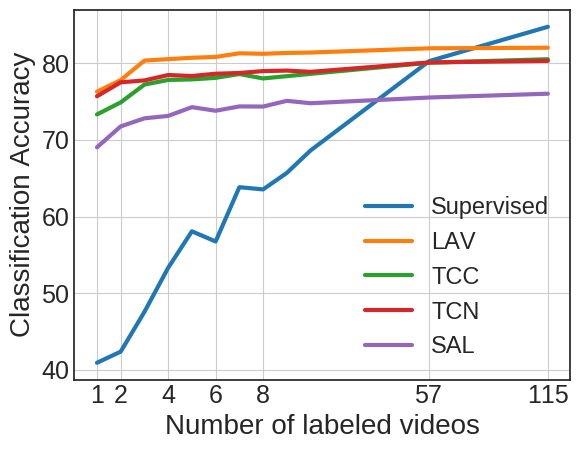}
        \caption{Tennis Serve}
        \label{fig:fs_tf}
    \end{subfigure}
    \vspace{-0.15cm}
    \caption{Few-shot phase classification results.}
    \label{fig:few_shot}
    \vspace{-0.15cm}
\end{minipage}
\end{figure}

In this section, we evaluate the utility of our representations for action phase classification. Tab.~\ref{tab:results_clf_sep} presents the quantitative results of all methods on \emph{Pouring}, \emph{Penn Action}, and \textit{IKEA ASM} datasets. For \emph{Penn Action} experiments, we follow~\cite{dwibedi2019temporal} to train 13 different models (i.e., 1 encoder + 1 SVM classifier, for each action) and report the average results across all actions. It can be seen from Tab.~\ref{tab:results_clf_sep} that our method (LAV) outperforms other self-supervised video representation learning methods, namely SAL~\cite{misra2016shuffle}, TCN~\cite{sermanet2018time}, and TCC~\cite{dwibedi2019temporal}, on all datasets. This shows that LAV is more capable of learning useful features that allow good classification performance when combined with a relatively simple classifier. Moreover, the best accuracy on \emph{Pouring} and \emph{IKEA ASM} is achieved by the combined LAV+TCC, which is similar to the observation in~\cite{dwibedi2019temporal}, where combining multiple losses leads to better classification performance. Next, the relative gaps between LAV and other self-supervised methods are the largest on \emph{IKEA ASM}, which has more complex actions than \emph{Pouring} and \emph{Penn Action}. This implies that LAV is more capable of handling complex actions. Finally, compared to the fully-supervised baseline, self-supervision with LAV provides a significant performance boost in the low labeled data regimes. Specifically, with just 10\% labeled data, LAV achieves very similar performance to the fully-supervised baseline trained with 100\% labeled data (e.g., on \emph{Penn Action}, $83.56\%$ vs. $84.34\%$).

\noindent \textbf{Few-Shot Phase Classification Results.} Following the above observation, we consider the application of our representations in a few-shot learning setting, i.e., there are many training videos, but only a few of them have frame-wise labels. We use the same setup as the above experiment, and compare our approach with other self-supervised methods and the fully-supervised baseline. For learning self-supervised features, all training videos are used, whereas the fully-supervised baseline is trained with a few labeled videos. Specifically, we study the classification performance with increasing the number of labeled videos. The results for two actions of \emph{Penn Action} are reported in Fig.~\ref{fig:few_shot}. Although all self-supervised methods offer a significant performance boost in the low labeled data settings, LAV provides the largest gain. Moreover, self-supervision using LAV with only 1 labeled video performs similarly to the fully-supervised baseline trained with the whole dataset. For instance, on \emph{Bowling}, with just 1 labeled video, LAV achieves 71\%, whereas the fully-supervised baseline trained with the entire dataset (134 labeled videos) obtains 77\%.

\subsection{Phase Progression and Kendall's Tau Results}
\label{sec:alignment}

\begin{table}
\vspace{-0.15cm}
\begin{minipage}[b]{1.0\linewidth}
\centering

{%

\begin{tabular}{l|l|c|c}

\specialrule{1pt}{1pt}{1pt}

 & \textbf{Method} & \textbf{Progress} & \bm{$\tau$} \\
\midrule

\multirow{5}{*}{\rotatebox[origin=c]{90}{\thead{\textbf{Pouring}}}}

& SAL~\cite{misra2016shuffle}
 
&   0.7728 & 0.7961
\\

& TCN~\cite{sermanet2018time}
&  0.8044 & 0.8521
  \\

& TCC~\cite{dwibedi2019temporal} 
 
&   \textbf{0.8373} & \textbf{0.8636}
 \\

& LAV (Ours)
 
&   \underline{0.8054} & \underline{0.8561}\\

& LAV + TCC (Ours)
 
&    0.7716 & 0.7844\\

\midrule
\multirow{5}{*}{\rotatebox[origin=c]{90}{\textbf{\thead{Penn  Action}}}}

& SAL~\cite{misra2016shuffle}
 
& 0.6960 &  0.7612
\\

& TCN~\cite{sermanet2018time}
 
&  \textbf{0.7217} & \textbf{0.8120}
\\

& TCC~\cite{dwibedi2019temporal} 
 
&  0.6638 & 0.7012 
\\

& LAV (Ours)
 
& 0.6613 &  \underline{0.8047}
\\

& LAV + TCC (Ours)
 
& \underline{0.7038} & 0.7729 
\\







\specialrule{1pt}{1pt}{1pt}

\end{tabular}

}

\caption{Phase progression and Kendall's Tau results. Best results are in \textbf{bold}, while second best ones are \underline{underlined}.}
\label{tab:results_kt_progress}
\vspace{-0.15cm}
\end{minipage}

\end{table}

We now evaluate the performance of our approach on action phase progression and Kendall's Tau. Tab.~\ref{tab:results_kt_progress} presents the quantitative results of different self-supervised methods on \emph{Pouring} and \emph{Penn Action}. We do not evaluate on \emph{IKEA ASM}, since its labels are repeated (i.e., the actions of picking up left side panel and picking up right side panel are both labeled as \emph{Pick Up Side Panel}, thus \emph{Pick Up Side Panel} is repeated). From the results, we achieve competitive numbers for both progression and Kendall's Tau on both \emph{Pouring} and \emph{Penn Action}. On \emph{Pouring}, LAV marginally beats TCN on both metrics (e.g., for progression, $0.8054$ vs. $0.8044$), while on \emph{Penn Action}, LAV significantly outperforms TCC on Kendall's Tau (i.e., $0.8047$ vs. $0.7012$). Moreover, on \emph{Penn Action}, the combination of LAV+TCC yields a significant performance gain over TCC on both metrics (e.g., for Kendall's Tau, $0.7729$ vs. $0.7012$).

\subsection{Fine-Grained Frame Retrieval Results}
\label{sec:retrival}

\begin{table}
\vspace{-0.15cm}
\begin{minipage}[b]{1.0\linewidth}
\centering

{%

\begin{tabular}{c|c|ccc}

\specialrule{1pt}{1pt}{1pt}
 & \textbf{Method}  & {\textbf{AP@5}} & {\textbf{AP@10}} & {\textbf{AP@15}} \\
\midrule
\multirow{4}{*}{\rotatebox[origin=c]{90}{\small{\textbf{Pouring}}}}
& SAL~\cite{misra2016shuffle}
 
&   84.05 & 83.77 & 83.79
\\

& TCN~\cite{sermanet2018time}
&  83.56 & 83.31 & 83.01
\\

& TCC~\cite{dwibedi2019temporal} 
 
&  87.16 & 86.68 & 86.54
 \\
 
& {LAV (Ours)} 
 
& \textbf{89.13} & \textbf{89.13} & \textbf{89.22} \\

& {LAV + TCC (Ours)} 
 
& \underline{89.00} & \underline{88.96} & \underline{88.78}

\\

\midrule

\multirow{4}{*}{\rotatebox[origin=c]{90}{\small{\textbf{Penn Action}}}}
& SAL~\cite{misra2016shuffle}
&  76.04 & 75.77 & 75.61 
\\

& TCN~\cite{sermanet2018time}
&   77.84 & 77.51 & 77.28
\\

& TCC~\cite{dwibedi2019temporal} 
 &   76.74 & 76.27 & 75.88
\\

& {LAV (Ours)}
&  \textbf{79.13} & \textbf{78.98} & \textbf{78.90}
\\

& {LAV + TCC (Ours)}
&   \underline{78.98} & \underline{78.83} & \underline{78.70}

\\

\specialrule{1pt}{1pt}{1pt}

 
\end{tabular}

}

\caption{Fine-grained frame retrieval results. Best results are in \textbf{bold}, while second best ones are \underline{underlined}.}
\label{tab:results_retrieval}
\vspace{-0.15cm}
\end{minipage}

\end{table}

\begin{figure}[t]
	\centering
		\includegraphics[width=1.0\linewidth, trim = 0mm 5mm 50mm 0mm, clip]{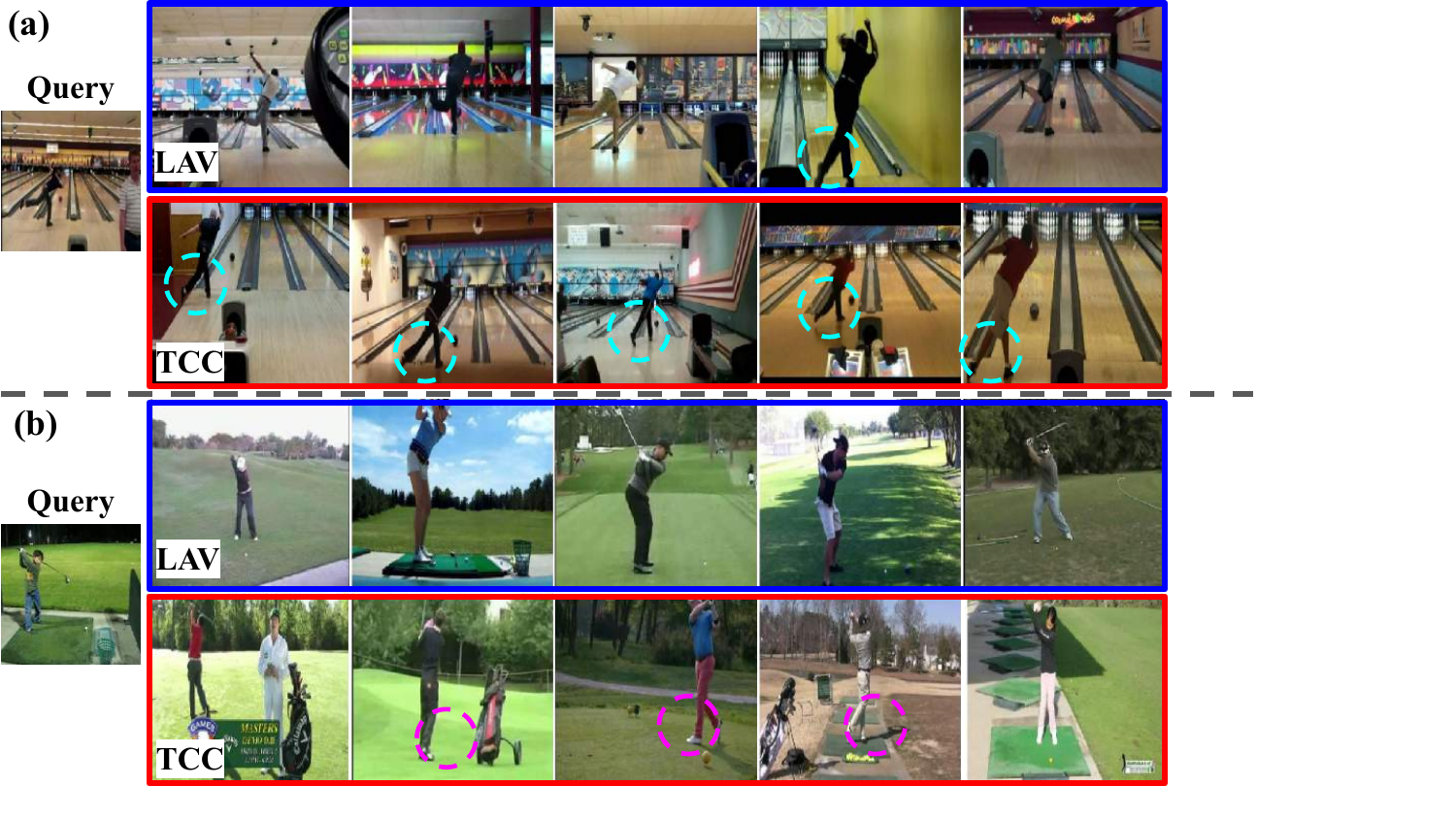}
    \vspace{-0.3cm}
	\caption{Qualitative fine-grained frame retrieval results with $K = 5$. On the left is the query image. On the right are the \textcolor{blue}{\textbf{blue}} and \textcolor{red}{\textbf{red}} boxes containing the 5 most similar images to the query image retrieved by LAV and TCC respectively.}
	\label{fig:ret_qual}
	\vspace{-0.2cm}
\end{figure}

Here, we utilize our representations for the task of fine-grained frame retrieval. We perform evaluations using the validation set of \emph{Pouring} and \emph{Penn Action}. In particular, we alternatively consider each video of the validation set as a query video and all the remaining videos of the validation set as a support set. For each query frame in the query video, we retrieve its $K$ most similar frames in the support set by finding its $K$ nearest neighbors in the embedding space. We report Average Precision at $K$, which is the average percentage of the $K$ retrieved frames with the same action phase labels as the query frame. Tab.~\ref{tab:results_retrieval} presents the quantitative results of various self-supervised methods on \emph{Pouring} and \emph{Penn Action}. It is evident from Tab.~\ref{tab:results_retrieval} that LAV consistently achieves the best performance across different values of $K$ on both datasets (e.g., on \emph{Pouring}, for AP@5, $89.13\%$ for LAV vs. $87.16\%$, $83.56\%$, and $84.05\%$ for TCC, TCN, and SAL respectively). This shows that our method is better at learning fine-grained features, which are important to this task. Also, the combined LAV+TCC leads to a significant performance gain over TCC (e.g., $78.7\%$ vs. $75.88\%$).

Moreover, we present some qualitative results with $K = 5$ in Fig.~\ref{fig:ret_qual}, showing that LAV is more capable of capturing fine-grained features than TCC. In Fig.~\ref{fig:ret_qual}(a), the person in the query image has one leg elevated above the ground, which is also seen in 4 out of 5 images retrieved by LAV, whereas TCC fails to capture that in all of its retrieved images (see cyan circles). In Fig.~\ref{fig:ret_qual}(b), the actor in the query image is at the start of \emph{Golf Swing} with the ball on the ground, which is also seen in all of LAV's retrieved images, whereas TCC retrieves images with wrong phases (i.e., the person has finished \emph{Golf Swing} with the ball not visible on the ground, see magenta circles).

\subsection{Joint All-Action Model Results}
\label{sec:joint} 

\begin{table}
\vspace{-0.15cm}
\begin{minipage}[b]{1.0\linewidth}
\centering

{%

\begin{tabular}{l|l|l|l|l}

\specialrule{1pt}{1pt}{1pt}

 & \textbf{\small{Method}} & \textbf{\small{Classification}} & \textbf{\small{Progress}} & \bm{$\tau$}\\
\midrule

\multirow{4}{*}{\rotatebox[origin=c]{90}{\small{\textbf{Penn Action}}}}
&SAL~\cite{misra2016shuffle}
 
&   68.15 & 0.3903 & 0.4744 

  \\

 &TCN~\cite{sermanet2018time}
 
&  68.09 & 0.3834 & 0.5417
\\

& TCC~\cite{dwibedi2019temporal}
 
&  \underline{74.39} & \underline{0.5914} & \underline{0.6408}
\\

& LAV (Ours)
 
&   \textbf{78.68} & \textbf{0.6252} & \textbf{0.6835}
\\

\specialrule{1pt}{1pt}{1pt}
\end{tabular}

}

\caption{Joint all-action model results. Best results are in \textbf{bold}, while second best ones are \underline{underlined}.}
\label{tab:results_clf_joint}
\vspace{-0.15cm}
\end{minipage}

\end{table}

So far, we have followed~\cite{dwibedi2019temporal} to train a separate model for each action of \emph{Penn Action} and report the average results across all actions. This is not convenient both in terms of training time and memory requirement. In this section, we explore another experimental setup, where we jointly train a single model for all actions of \emph{Penn Action}. In particular, we train 13 SVM classifiers (1 for each action) but share a single encoder. It is more challenging, since the network needs to jointly learn useful features for all actions. Tab.~\ref{tab:results_clf_joint} shows the quantitative results of different self-supervised methods in the above setup. We observe that the performance of all methods is reduced as compared to Tabs.~\ref{tab:results_clf_sep} and~\ref{tab:results_kt_progress}. Moreover, we notice LAV achieves the best performance across all metrics, outperforming TCC, TCN, and SAL in Tab.~\ref{tab:results_clf_joint}. This can be attributed to the fact that LAV leverages information from across videos in addition to cues from each individual video.

\noindent \textbf{Additional Results.} Note that due to space limits, we provide several additional experimental results, including training-from-scratch results and ablation results of hyperparameter settings, in supplementary materials.
\section{Conclusion}
\label{sec:conclusion}

In this work, we propose a novel fusion of temporal alignment loss and temporal regularization for learning self-supervised video representations via temporal video alignment, utilizing both frame-level and video-level cues. The two components are complementary to each other, i.e., temporal regularization prevents degenerate solutions while temporal alignment loss leads to higher performance. We show superior performance over prior methods for video-based self-supervised representation learning on various temporal understanding tasks on \emph{Pouring}, \emph{Penn Action}, and \emph{IKEA ASM} datasets. Also, our method offers significant accuracy gain when lacking labeled data. Our future work will explore other temporal alignment losses, e.g.,~\cite{su2017order,cao2020few}, to allow local temporal permutations and arbitrary video starting/ending points.

\noindent \textbf{Acknowledgements.} We would like to thank D. Dwibedi for releasing the code  and answering questions about TCC.

\appendix
\section{Supplementary Material}

In this supplementary material, we first present results on a subset of 11 actions of \emph{Penn Action} in Sec.~\ref{sec:supp-11actions} and fine-grained frame retrieval results on \emph{IKEA ASM} in Sec.~\ref{sec:supp-retrival}. We then show training-from-scratch results in Sec.~\ref{sec:supp-scratch} and ablation results of $\alpha$, $\sigma$, and $p$ in Sec.~\ref{sec:supp-ablation}. Next, in Sec.~\ref{sec:supp-lav+tcn} we show results of combining LAV with TCC and TCN while we show results on a recent frame-shuffling method in Sec.~\ref{sec:supp-recent-shuffling}. Moreover, we visualize our embeddings and provide our labels for \emph{Penn Action} in Secs.~\ref{sec:supp-visualization} and~\ref{sec:supp-annotation} respectively. Finally, we describe our additional implementation details in Sec.~\ref{sec:supp-implementation}.

\subsection{Results on a Subset of 11 Actions of \emph{Penn Action}}
\label{sec:supp-11actions}

\begin{figure}
\begin{minipage}[b]{1.0\linewidth}
    \centering
    \includegraphics[width=0.9\linewidth]{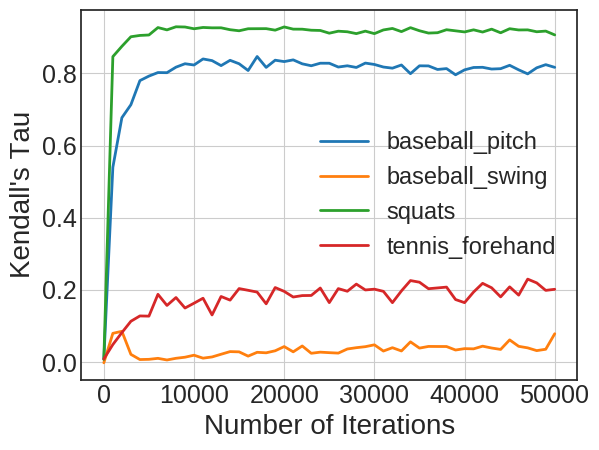}
    \caption{Kendall's Tau results by TCC.}
    \label{fig:tcc_bstf}
\end{minipage}
\end{figure}

\begin{figure}[t]
     \centering
     \begin{subfigure}[b]{0.45\textwidth}
         \centering
         \includegraphics[width=\textwidth]{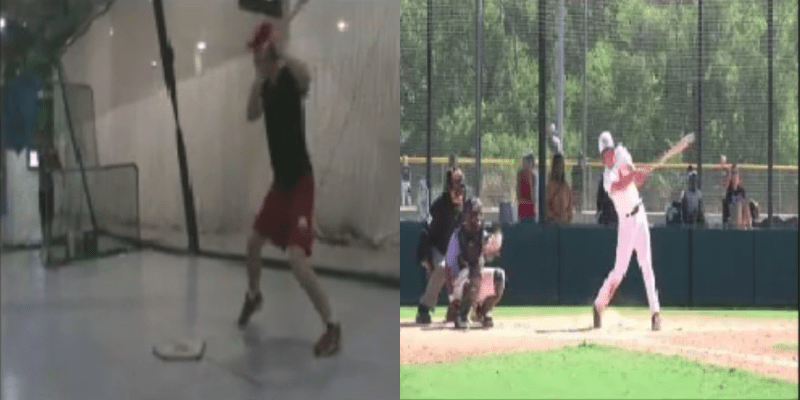}
         \caption{Baseball Swing}
         \label{fig:tsne_bs}
     \end{subfigure}
     \\
     \begin{subfigure}[b]{0.45\textwidth}
         \centering
         \includegraphics[width=\textwidth]{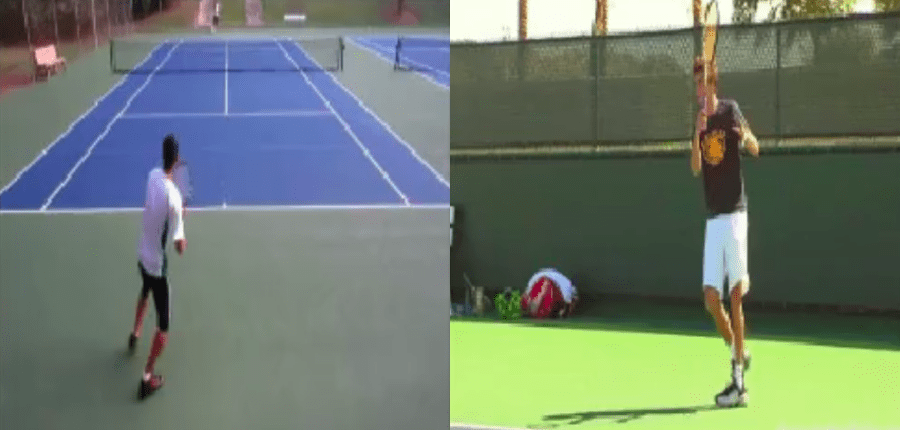}
         \caption{Tennis Forehand}
         \label{fig:tsne_push}
     \end{subfigure}
     \\
    \caption{Example alignment errors by TCC. On the left is the reference frame in one video, and on the right is the aligned frame in another video by TCC (i.e., via nearest neighbor search in the embedding space). The frame on the left is among the beginning frames, while the frame on the right in among the ending frames. TCC incorrectly aligns them together due to their similar appearances/poses.}
    \label{fig:tcc_qual}
\end{figure}
\begin{table}

\begin{minipage}[b]{1.0\linewidth}
\centering

{%
\begin{tabular}{l|l|l|l|l}
\specialrule{1pt}{1pt}{1pt}

 & \textbf{\small{Method}} & \textbf{\small{Classification}} & \textbf{\small{Progress}} & \bm{$\tau$}\\
\midrule

\multirow{4}{*}{\rotatebox[origin=c]{90}{\small{\textbf{Penn Action}}}} 

& SAL~\cite{misra2016shuffle} & 82.03 & 0.7054 & 0.7783
\\

& TCN~\cite{sermanet2018time} & 82.99 & \underline{0.7281} & \textbf{0.8275}
\\

& TCC~\cite{dwibedi2019temporal} & 83.94 & \textbf{0.7394} & 0.8001
\\

& LAV (Ours) & \textbf{84.47} & 0.6654 & \underline{0.8149}
\\

& LAV+TCC (Ours) & \underline{84.14} & 0.7111 & 0.7892
\\

\specialrule{1pt}{1pt}{1pt}
\end{tabular}
}
\caption{Phase classification, phase progression, and Kendall's Tau results on a subset of 11 actions of \emph{Penn Action}. Best results are in \textbf{bold}, while second best ones are \underline{underlined}.}
\label{tab:penn_action_11act}

\end{minipage}

\end{table}

\begin{table}[h!]

\begin{minipage}[b]{1.0\linewidth}
\centering

{%

\begin{tabular}{c|c|ccc}

\specialrule{1pt}{1pt}{1pt}
 & \textbf{Method}  & {\textbf{AP@5}} & {\textbf{AP@10}} & {\textbf{AP@15}} \\
\midrule
\multirow{4}{*}{\rotatebox[origin=c]{90}{\small{\textbf{Penn Action}}}}

& SAL~\cite{misra2016shuffle} & 76.83 & 76.52 & 76.35
\\

& TCN~\cite{sermanet2018time} & 78.09 & 77.74 & 77.52
\\

& TCC~\cite{dwibedi2019temporal} & \textbf{79.49} & \underline{79.19} & \underline{79.00}
\\
 
& {LAV (Ours)} &  \underline{79.34} & \textbf{79.23} & \textbf{79.08 }
\\

& {LAV + TCC (Ours)} &  79.29  &  79.15  &  79.03
\\

\specialrule{1pt}{1pt}{1pt}
\end{tabular}

}

\caption{Fine-grained frame retrieval results on a subset of 11 actions of \emph{Penn Action}. Best results are in \textbf{bold}, while second best ones are \underline{underlined}.}
\label{tab:retrieval_pa_11actions}

\end{minipage}

\end{table}

Among the 3 datasets that we use in Sec.~5 of the main paper (i.e., \emph{Pouring}, \emph{Penn Action}, and \emph{IKEA ASM}), we notice that for \emph{Penn Action} while TCC performs well on most actions, it struggles on 2 actions i.e., \emph{Baseball Swing} and \emph{Tennis Forehand}. As we can see in Fig.~\ref{fig:tcc_bstf}, the Kendall's Tau results by TCC on the above 2 actions (red and orange curves) do not go higher than $\sim 0.2$, which hurts its overall performance on \emph{Penn Action} in Tabs.~2-4 of the main paper. This might be due to the fact that the beginning and ending frames of the above 2 actions are visually similar, and TCC does not have an explicit mechanism to avoid aligning the beginning frames with the ending ones and vice versa (see Fig.~\ref{fig:tcc_qual} for examples). Other self-supervised methods, i.e., SAL, TCN, and LAV, do not suffer from the above problem as TCC, since they leverage temporal order information, i.e., SAL performs temporal order verification, TCN uses temporal coherence, while LAV exploits both temporal coherence and dynamic time warping prior. Also, the above problem for TCC might be alleviated by tuning the number of context frames and context stride, however, that requires further exploration.

For completeness, we filter out the results of the above 2 actions from Tabs.~2-4 of the main paper, and present the results of the remaining 11 actions of \emph{Penn Action} in Tabs.~\ref{tab:penn_action_11act} and~\ref{tab:retrieval_pa_11actions}. From the results, the performance of TCC is improved significantly when excluding the above 2 actions. In Tab.~\ref{tab:penn_action_11act}, TCC has competitive numbers with TCN and LAV (e.g., TCC performs the best on progression, while TCN and LAV perform the best on Kendall's Tau and classification respectively). In Tab.~\ref{tab:retrieval_pa_11actions}, TCC and LAV have very competitive numbers (e.g., TCC slightly outperforms LAV for AP@5, while LAV marginally outperforms TCC for AP@10 and AP@15), outperforming SAL and TCN.

\subsection{Fine-Grained Frame Retrieval Results on \emph{IKEA ASM}}
\label{sec:supp-retrival}

We now conduct fine-grained frame retrieval experiments on \emph{IKEA ASM} and report the quantitative results of different self-supervised methods in Tab.~\ref{tab:ASM_results_retrieval}. It is evident from the results that LAV consistently achieves the best performance across different values of $K$, outperforming other methods by significant margins. For example, for AP@5, LAV obtains $23.89\%$, while TCC, TCN, and SAL get $19.80$\%, $19.15$\%, and $15.15$\% respectively. Furthermore, the combined LAV+TCC leads to significant performance increase over TCC. For instance, for AP@5, LAV+TCC achieves $22.95\%$, while TCC obtains $19.80\%$. The above observations on \emph{IKEA ASM} are similar to those on \emph{Penn Action} and \emph{Pouring} reported in Sec.~5.4 of the main paper, confirming the utility of our self-supervised representation for fine-grained frame retrieval.

\begin{table}[t]

\begin{minipage}[b]{1.0\linewidth}
\centering

{%

\begin{tabular}{c|c|ccc}

\specialrule{1pt}{1pt}{1pt}
 & \textbf{Method}  & {\textbf{AP@5}} & {\textbf{AP@10}} & {\textbf{AP@15}} \\
\midrule
\multirow{4}{*}{\rotatebox[origin=c]{90}{\small{\textbf{IKEA ASM}}}}
& SAL~\cite{misra2016shuffle}
 
&    15.15 & 14.90 & 14.72

\\

& TCN~\cite{sermanet2018time}
&   19.15  & 19.19 & 19.33

\\

& TCC~\cite{dwibedi2019temporal} 
 
&   19.80 & 19.64 & 19.68

\\
 
& {LAV (Ours)} 
 
&  \textbf{23.89} & \textbf{23.65} & \textbf{23.56} \\

& {LAV + TCC (Ours)} 
 
&  \underline{22.95} & \underline{22.80} & \underline{22.70} \\

\specialrule{1pt}{1pt}{1pt}

 
\end{tabular}

}

\caption{Fine-grained frame retrieval results on \emph{IKEA ASM}. Best results are in \textbf{bold}, while second best ones are \underline{underlined}.}
\label{tab:ASM_results_retrieval}

\end{minipage}

\end{table}

\subsection{Training-from-Scratch Results}
\label{sec:supp-scratch}

\begin{table}
\vspace{-0.15cm}
\begin{minipage}[b]{1.0\linewidth}
\centering

{%

\begin{tabular}{l|l|l|l|l}

\specialrule{1pt}{1pt}{1pt}

 & \textbf{\small{Method}} & \textbf{\small{Classification}} & \textbf{\small{Progress}} & \bm{$\tau$}\\
\midrule

\multirow{4}{*}{\rotatebox[origin=c]{90}{\small{\textbf{Pouring}}}}
&SAL~\cite{misra2016shuffle}
 
& 85.86 & 0.6422 & 0.7329
 
 \\

&TCN~\cite{sermanet2018time}
 
&   85.98 & 0.6732 & 0.7500

\\

& TCC~\cite{dwibedi2019temporal}
 
&   \textbf{88.59} & \underline{0.7104} & \underline{0.7774}

\\

& LAV (Ours)
 
&  \underline{87.70} & \textbf{0.7320} & \textbf{0.7867}

\\

\specialrule{1pt}{1pt}{1pt}
\end{tabular}

}

\caption{Training-from-scratch results on \emph{Pouring}. Best results are in \textbf{bold}, while second best ones are \underline{underlined}.}
\label{tab:results_scratch}
\vspace{-0.15cm}
\end{minipage}

\end{table}

All of the experiments in Sec. 5 of the main paper utilize an encoder network initialized with pre-trained weights from ImageNet classification. For completeness, we now experiment with learning from scratch. We use a smaller backbone network, i.e., VGG-M~\cite{chatfield2014return}, (instead of ResNet-50) for this experiment. Tab.~\ref{tab:results_scratch} shows the quantitative results of different self-supervised methods when learning from scratch on \emph{Pouring}. It can be seen from Tab.~\ref{tab:results_scratch} that the performance of all methods drops as compared to Tabs.~2 and~3 of the main text. Moreover, SAL and TCN are inferior to TCC and LAV across all metrics. Lastly, although LAV has slightly lower classification accuracy than TCC, LAV outperforms TCC on both progression and Kendall's Tau.

Next, Tab.~\ref{tab:scratch_joint} shows training-from-scratch results on \emph{Penn Action}, using a single joint model for all actions (similar as Sec.~5.5 of the main paper). For all methods, the performance in Tab.~\ref{tab:scratch_joint} is lower than Tab.~5 of the main text. Also, SAL and TCN are inferior to TCC and LAV. TCC performs the best on progression, while LAV performs the best on the other two metrics.

Finally, we obtain training-from-scratch results on \emph{IKEA ASM}, which show LAV achieves the best performance (i.e., for classification, \textbf{23.84} for LAV vs. \textbf{22.04}, \textbf{20.45}, and \textbf{20.42} for TCC, TCN, and SAL respectively).

\begin{table}

\begin{minipage}[b]{1.0\linewidth}
\centering

{%

\begin{tabular}{l|l|l|l|l}

\specialrule{1pt}{1pt}{1pt}

 & \textbf{\small{Method}} & \textbf{\small{Classification}} & \textbf{\small{Progress}} & \bm{$\tau$}\\
\midrule

\multirow{4}{*}{\rotatebox[origin=c]{90}{\small{\textbf{Penn Action}}}}
&SAL 
 
& 64.05 & 0.2989 & 0.4145
 
 \\

&TCN 
 
& 60.17 & 0.1909 & 0.4260

\\

& TCC 
 
& \underline{65.53} & \textbf{0.4304} & \underline{0.4529}

\\

& LAV (Ours)
 
& \textbf{67.90} & \underline{0.3853} & \textbf{0.4929}

\\

\specialrule{1pt}{1pt}{1pt}
\end{tabular}

}

\caption{Training-from-scratch results on \emph{Penn Action}. Best results are in \textbf{bold}. Second best results are \underline{underlined}.}
\label{tab:scratch_joint}

\end{minipage}

\end{table}

\subsection{Ablation Results of $\alpha$, $\sigma$, and $p$}
\label{sec:supp-ablation}

\begin{figure}
\begin{minipage}[b]{1.0\linewidth}
    \centering
    \begin{subfigure}[t]{0.32\textwidth}
        \centering
        \includegraphics[height=0.83in]{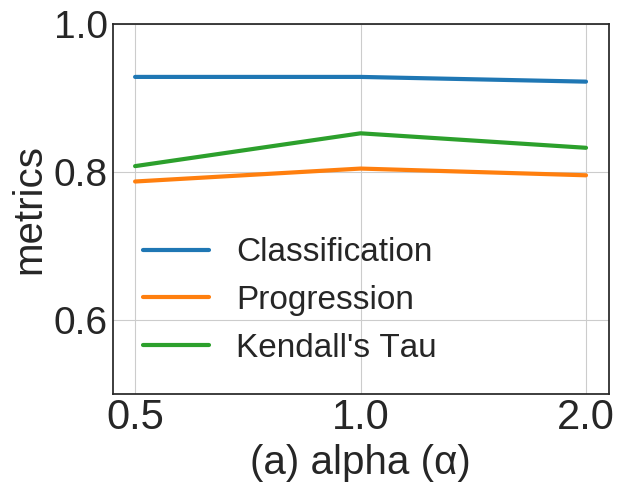}
        \label{fig:fs_bowl}
    \end{subfigure}%
    \begin{subfigure}[t]{0.32\textwidth}
        \centering
        \includegraphics[height=0.83in]{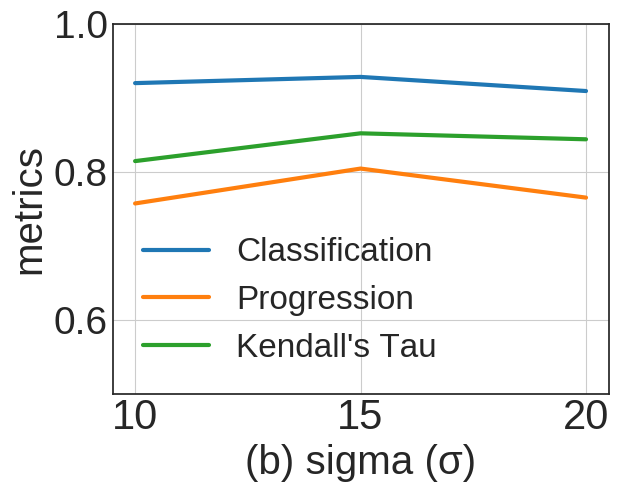}
        \label{fig:fs_tf}
    \end{subfigure}
    \begin{subfigure}[t]{0.32\textwidth}
        \centering
        \includegraphics[height=0.83in]{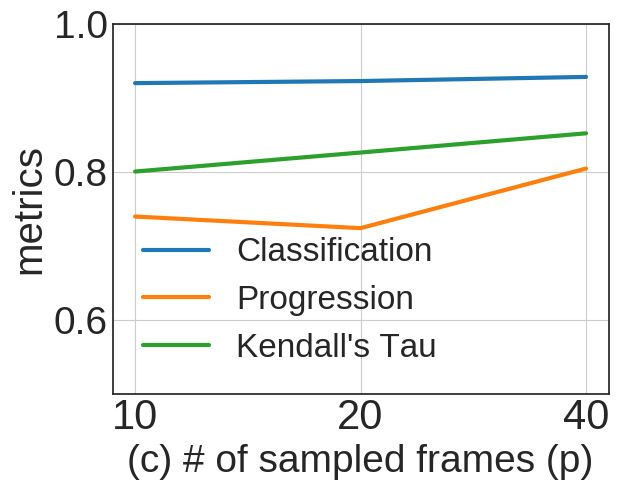}
        \label{fig:fs_tf}
    \end{subfigure}

    \caption{Ablation results of $\alpha$ in (a), $\sigma$ in (b), and $p$ in (c) on \emph{Pouring}. We convert classification results from \% to $[0,1]$.}
    \label{fig:ablation_extra}
    
\end{minipage}
\end{figure}

We first present ablation results of $\alpha$ on \emph{Pouring} in Fig.~\ref{fig:ablation_extra}(a). We observe that the performance is generally stable across values of $\alpha$, and $\alpha=1.0$ yields the best results. Next,  Figs.~\ref{fig:ablation_extra}(b) and~\ref{fig:ablation_extra}(c) illustrate ablation results of $\sigma$ and $p$ respectively on \emph{Pouring}. From the results, the performance is generally stable across values of $\sigma$ and $p$. Particularly, $\sigma=15$ performs the best, and large $p$ is preferred.

\subsection{Performance of LAV+TCC and LAV+TCN}
\label{sec:supp-lav+tcn}

We note that LAV+TCC does not consistently perform better than LAV in Tabs.~2 and~3 of the main paper. This might be attributed to the fact that LAV works on L2-normalized embeddings while TCC does not. Since the two components operate on different embedding spaces, combining the two might not always lead to better results.

In addition, we evaluate LAV+TCN on \emph{Pouring}. We notice LAV+TCN suffers from the same problem as LAV+TCC (i.e., normalized/unnormalized embeddings). LAV+TCN obtains \textbf{91.22}, \textbf{0.7866}, and \textbf{0.7925} for classification, progression, and Kendall's Tau respectively, which are comparable to TCN but lower than LAV.

\subsection{Performance of a Recent Frame-Shuffling Method}
\label{sec:supp-recent-shuffling}

We evaluate the clip order prediction (COP) method of Xu et. al.~\cite{xu2019self} on \emph{Pouring}. As it is a clip-based method, we use sliding windows to generate embeddings for frames at window centers. As mentioned in Sec.~4 of the main paper, the network is first trained for the pretext task and then frozen while we train SVM classifier/linear regressor for the main tasks. It achieves \textbf{79.44}, \textbf{0.5309}, and \textbf{0.6656} for classification, progression, and Kendall’s Tau respectively, which are lower than SAL in Tabs.~2 and~3 of the main paper. This is likely because the pretext task (i.e., COP) is clip-based, whereas the main tasks are frame-based and require capturing fine-grained frame-based details. Further, since we freeze the network while training SVM classifier/linear regressor, it could not disregard irrelevant clip-based details to focus on the one frame that matters.

\subsection{Visualization of Embeddings}
\label{sec:supp-visualization}

\begin{figure*}[hb]
     \centering
     \begin{subfigure}[b]{0.6\textwidth}
         \centering
         \includegraphics[width=\textwidth, trim = 0mm 20mm 5mm 0mm, clip]{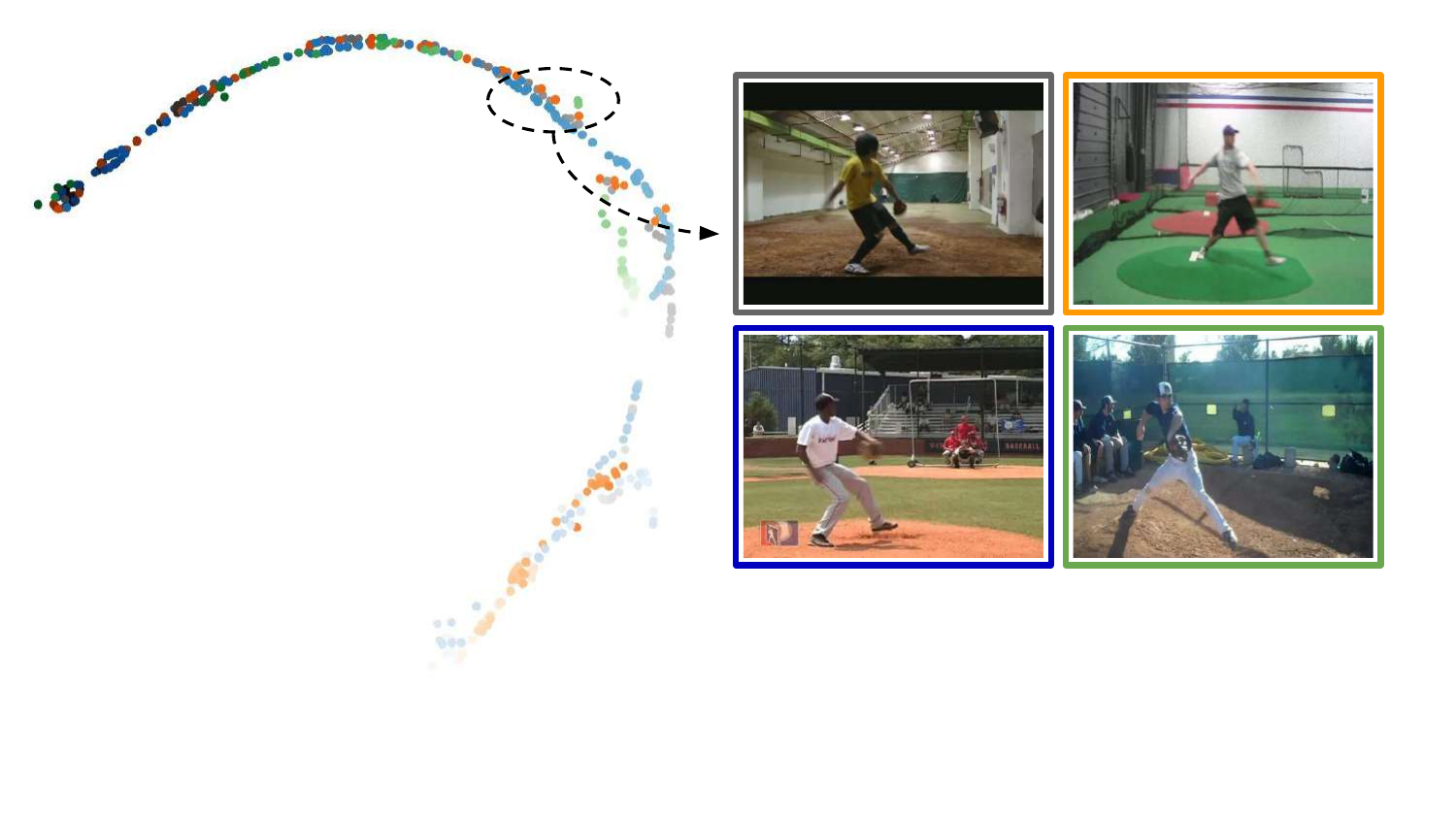}
         \caption{Baseball Pitch}
         \label{fig:tsne_bs}
     \end{subfigure}
     \\
     \begin{subfigure}[b]{0.6\textwidth}
         \centering
         \includegraphics[width=\textwidth, trim = 0mm 20mm 5mm 0mm, clip]{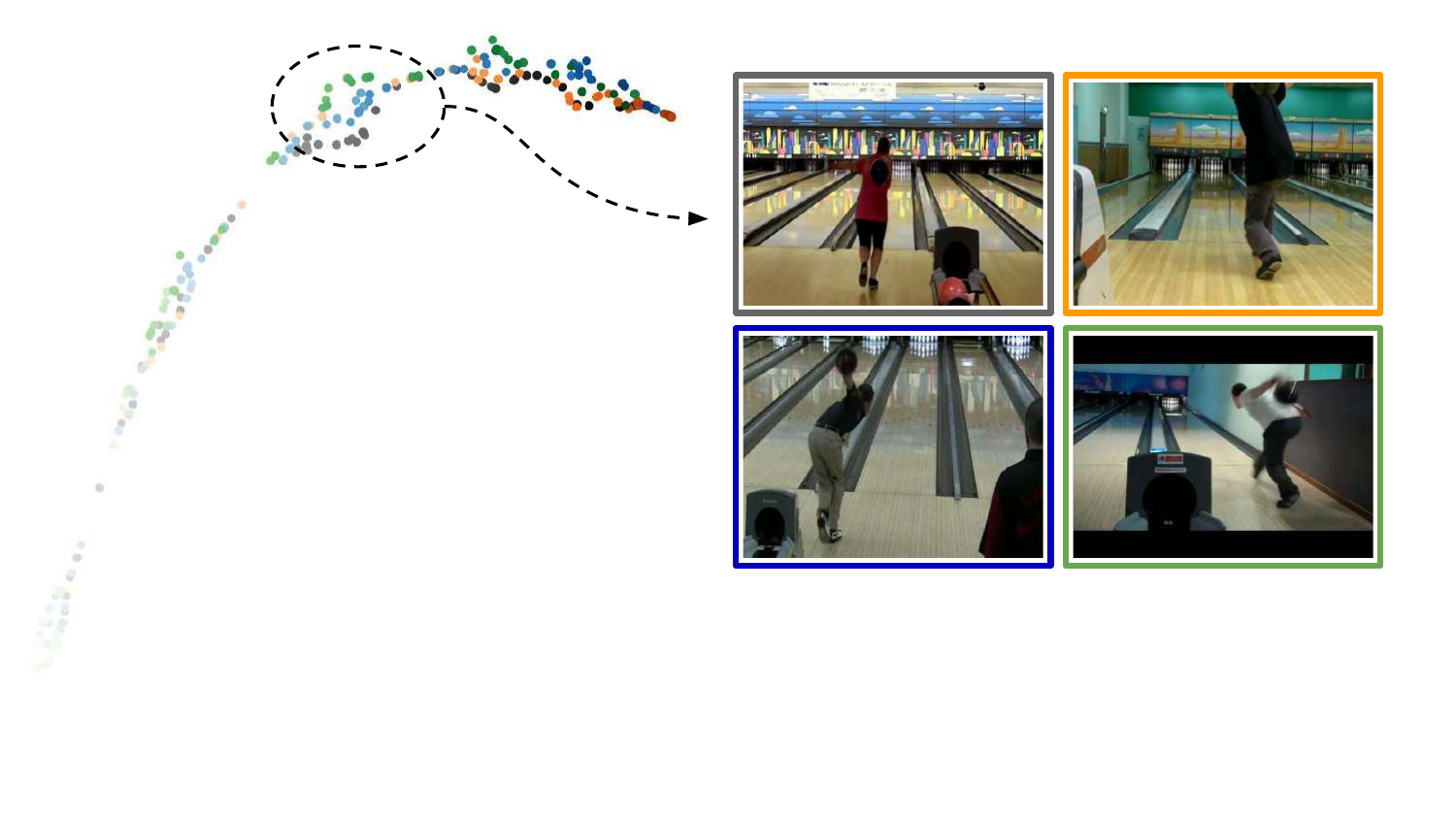}
         \caption{Bowling}
         \label{fig:tsne_push}
     \end{subfigure}
     \\
     \begin{subfigure}[b]{0.6\textwidth}
         \centering
         \includegraphics[width=\textwidth, trim = 0mm 20mm 5mm 0mm, clip]{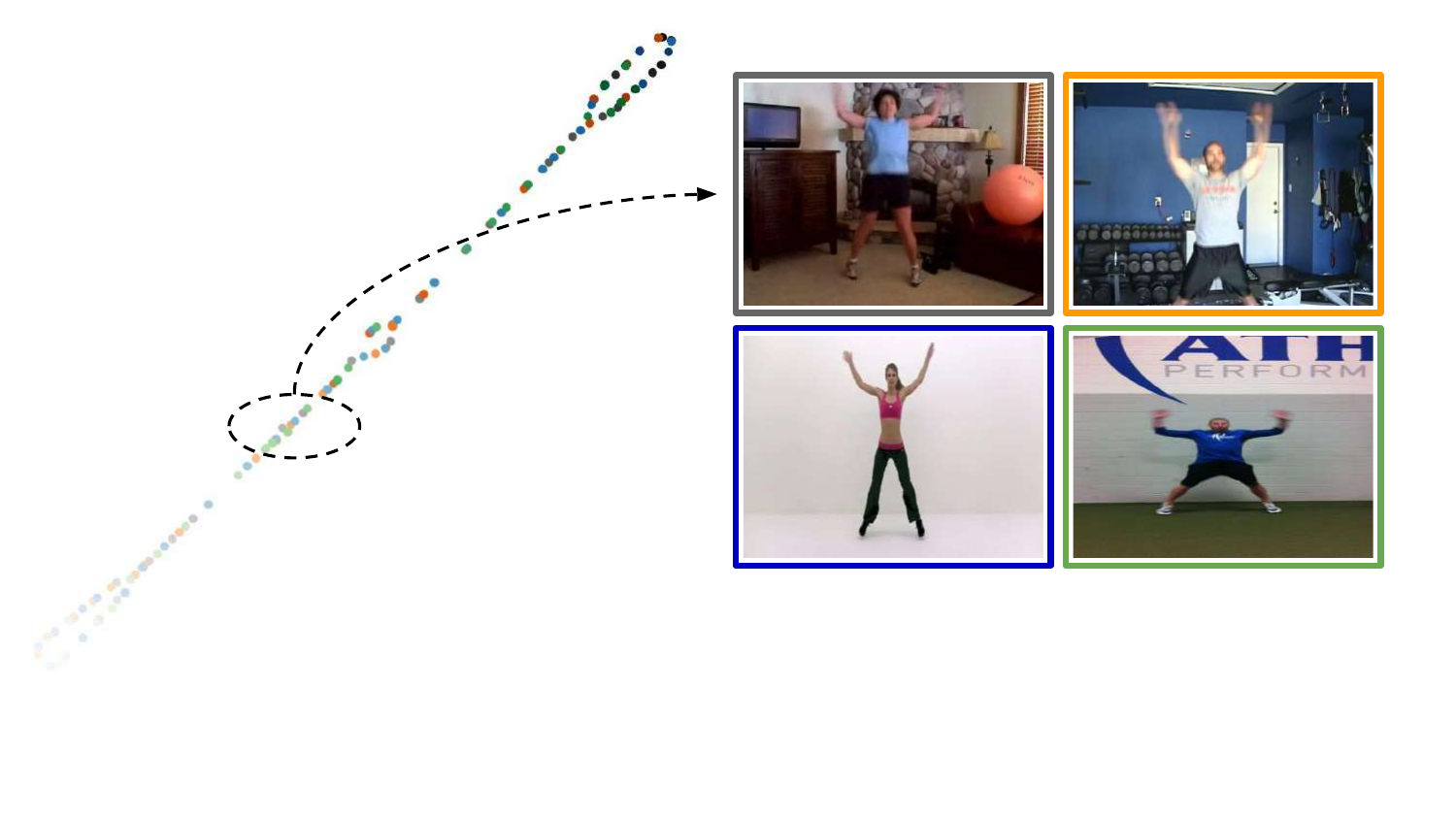}
         \caption{Jumping}
         \label{fig:tsne_pul}
     \end{subfigure}
    \caption{Visualization of embeddings for LAV.}
    \label{fig:TSNE}
\end{figure*}

We present the t-SNE visualization~\cite{maaten2008visualizing} of the embeddings learned by LAV on 3 example actions of \emph{Penn Action} in Fig.~\ref{fig:TSNE}. For each action, we show 4 videos with each plotted using a unique color. In addition, we use different shades of the same color to distinguish different frames of the same video, i.e., beginning frames have light shades, while later frames have progressively darker shades. The visualization in Fig.~\ref{fig:TSNE} shows that LAV encodes each video as an overall smooth trajectory in the embedding space, where temporally close frames are mapped to nearby points in the embedding space and vice versa. Moreover, corresponding frames from different videos are generally aligned in the embedding space, e.g., points of different colors but similar shades are nearby in the embedding space and vice versa. We also sample one random time-step (highlighted by a black circle), and plot corresponding frames from different videos (each bordered by a distinct color), which are shown to belong to the same action phase. The above observations show the potential application of our self-supervised representation for temporal video alignment.

\subsection{Labels for \emph{Penn Action}}
\label{sec:supp-annotation}

We have made our dense per-frame labels for 2123 videos of \emph{Penn Action} publicly available at \url{https://bit.ly/3f73e2W}. Please refer to Tab. 2 of TCC for more details on actions, numbers of phases, lists of key events, and numbers of videos for training and validation.

\subsection{Implementation Details}
\label{sec:supp-implementation}

\begin{table}

\begin{minipage}[b]{1.0\linewidth}
\centering

{%
\begin{tabular*}{1.0\linewidth}{L{0.45\linewidth}|R{0.45\linewidth}}

\specialrule{1pt}{1pt}{1pt}
\textbf{Hyperparameter} & \textbf{Value} \\
\midrule

\# of sampled frames ($p$) & 40 (P), 20 (PA, IA)
\\

Batch size & 1 (P), 2 (PA, IA)
\\

Learning rate & $10^{-4}$
\\

Weight decay & $10^{-5}$
\\

Soft-DTW smoothness ($\gamma$) & 0.1 
\\

Window size ($\sigma$) & 15 (P, IA), 7 (PA)
\\

Margin ($\lambda$) & 2
\\

Regularization weight ($\alpha$) & 1.0 (P), 0.5 (PA, IA)
\\

\# of context frames ($k$) & 1
\\

Context stride & 15 (P, PA), 8 (IA)
\\

\specialrule{1pt}{1pt}{1pt}
\end{tabular*}
}

\caption{Hyperparameter settings for LAV. Here, P denotes \emph{Pouring}, PA represents \emph{Penn Action}, and IA denotes \emph{IKEA ASM}. For batch size, 1 means 1 pair of videos (or 2 videos per batch).}
\label{tab:hyperparams}

\end{minipage}

\end{table}

For fair evaluations, we use the same data augmentation techniques and encoder networks for all the competing methods. More specifically, we follow the same data augmentation procedures and borrow the encoder networks from TCC~\cite{dwibedi2019temporal}. Please refer to the supplementary material of TCC for more details on data augmentation techniques and encoder networks. In addition, we list the hyperparameter settings for our method in Tab.~\ref{tab:hyperparams}. For other methods, we use the same hyperparameter settings suggested by TCC.

{\small
\bibliographystyle{ieee_fullname}
\bibliography{references}

\begin{thebibliography}{10}\itemsep=-1pt

\bibitem{ahsan2018discrimnet}
Unaiza Ahsan, Chen Sun, and Irfan Essa.
\newblock Discrimnet: Semi-supervised action recognition from videos using
  generative adversarial networks.
\newblock {\em arXiv preprint arXiv:1801.07230}, 2018.

\bibitem{ben2020ikea}
Yizhak Ben-Shabat, Xin Yu, Fatemeh~Sadat Saleh, Dylan Campbell, Cristian
  Rodriguez-Opazo, Hongdong Li, and Stephen Gould.
\newblock The ikea asm dataset: Understanding people assembling furniture
  through actions, objects and pose.
\newblock {\em arXiv preprint arXiv:2007.00394}, 2020.

\bibitem{bengio2009slow}
Yoshua Bengio and James~S Bergstra.
\newblock Slow, decorrelated features for pretraining complex cell-like
  networks.
\newblock In {\em Advances in neural information processing systems}, pages
  99--107, 2009.

\bibitem{berndt1994using}
Donald~J Berndt and James Clifford.
\newblock Using dynamic time warping to find patterns in time series.
\newblock In {\em KDD workshop}, volume~10, pages 359--370. Seattle, WA, USA:,
  1994.

\bibitem{cao2020few}
Kaidi Cao, Jingwei Ji, Zhangjie Cao, Chien-Yi Chang, and Juan~Carlos Niebles.
\newblock Few-shot video classification via temporal alignment.
\newblock In {\em Proceedings of the IEEE/CVF Conference on Computer Vision and
  Pattern Recognition}, pages 10618--10627, 2020.

\bibitem{carlucci2019domain}
Fabio~M Carlucci, Antonio D'Innocente, Silvia Bucci, Barbara Caputo, and
  Tatiana Tommasi.
\newblock Domain generalization by solving jigsaw puzzles.
\newblock In {\em Proceedings of the IEEE Conference on Computer Vision and
  Pattern Recognition}, pages 2229--2238, 2019.

\bibitem{carreira2017quo}
Joao Carreira and Andrew Zisserman.
\newblock Quo vadis, action recognition? a new model and the kinetics dataset.
\newblock In {\em proceedings of the IEEE Conference on Computer Vision and
  Pattern Recognition}, pages 6299--6308, 2017.

\bibitem{chang2019d3tw}
Chien-Yi Chang, De-An Huang, Yanan Sui, Li Fei-Fei, and Juan~Carlos Niebles.
\newblock {D3TW: Discriminative differentiable dynamic time warping for weakly
  supervised action alignment and segmentation}.
\newblock In {\em Proceedings of the IEEE Conference on Computer Vision and
  Pattern Recognition}, pages 3546--3555, 2019.

\bibitem{chatfield2014return}
Ken Chatfield, Karen Simonyan, Andrea Vedaldi, and Andrew Zisserman.
\newblock Return of the devil in the details: Delving deep into convolutional
  nets.
\newblock {\em arXiv preprint arXiv:1405.3531}, 2014.

\bibitem{choi2020shuffle}
Jinwoo Choi, Gaurav Sharma, Samuel Schulter, and Jia-Bin Huang.
\newblock Shuffle and attend: Video domain adaptation.
\newblock In {\em European Conference on Computer Vision}, pages 678--695.
  Springer, 2020.

\bibitem{conners1980theoretical}
Richard~W Conners and Charles~A Harlow.
\newblock A theoretical comparison of texture algorithms.
\newblock {\em IEEE transactions on pattern analysis and machine intelligence},
  (3):204--222, 1980.

\bibitem{cuturi2017soft}
Marco Cuturi and Mathieu Blondel.
\newblock Soft-dtw: a differentiable loss function for time-series.
\newblock In {\em International Conference on Machine Learning}, pages
  894--903, 2017.

\bibitem{diba2019dynamonet}
Ali Diba, Vivek Sharma, Luc~Van Gool, and Rainer Stiefelhagen.
\newblock Dynamonet: Dynamic action and motion network.
\newblock In {\em Proceedings of the IEEE International Conference on Computer
  Vision}, pages 6192--6201, 2019.

\bibitem{draper1998applied}
Norman~R Draper and Harry Smith.
\newblock {\em Applied regression analysis}, volume 326.
\newblock John Wiley \& Sons, 1998.

\bibitem{dwibedi2019temporal}
Debidatta Dwibedi, Yusuf Aytar, Jonathan Tompson, Pierre Sermanet, and Andrew
  Zisserman.
\newblock Temporal cycle-consistency learning.
\newblock In {\em Proceedings of the IEEE Conference on Computer Vision and
  Pattern Recognition}, pages 1801--1810, 2019.

\bibitem{farha2019ms}
Yazan~Abu Farha and Jurgen Gall.
\newblock Ms-tcn: Multi-stage temporal convolutional network for action
  segmentation.
\newblock In {\em Proceedings of the IEEE Conference on Computer Vision and
  Pattern Recognition}, pages 3575--3584, 2019.

\bibitem{feng2019self}
Zeyu Feng, Chang Xu, and Dacheng Tao.
\newblock Self-supervised representation learning by rotation feature
  decoupling.
\newblock In {\em Proceedings of the IEEE Conference on Computer Vision and
  Pattern Recognition}, pages 10364--10374, 2019.

\bibitem{fernando2017self}
Basura Fernando, Hakan Bilen, Efstratios Gavves, and Stephen Gould.
\newblock Self-supervised video representation learning with odd-one-out
  networks.
\newblock In {\em Proceedings of the IEEE conference on computer vision and
  pattern recognition}, pages 3636--3645, 2017.

\bibitem{gammulle2019predicting}
Harshala Gammulle, Simon Denman, Sridha Sridharan, and Clinton Fookes.
\newblock Predicting the future: A jointly learnt model for action
  anticipation.
\newblock In {\em Proceedings of the IEEE International Conference on Computer
  Vision}, pages 5562--5571, 2019.

\bibitem{gidaris2018unsupervised}
Spyros Gidaris, Praveer Singh, and Nikos Komodakis.
\newblock Unsupervised representation learning by predicting image rotations.
\newblock In {\em International Conference on Learning Representations}, 2018.

\bibitem{goroshin2015unsupervised}
Ross Goroshin, Joan Bruna, Jonathan Tompson, David Eigen, and Yann LeCun.
\newblock Unsupervised learning of spatiotemporally coherent metrics.
\newblock In {\em Proceedings of the IEEE international conference on computer
  vision}, pages 4086--4093, 2015.

\bibitem{hadsell2006dimensionality}
Raia Hadsell, Sumit Chopra, and Yann LeCun.
\newblock Dimensionality reduction by learning an invariant mapping.
\newblock In {\em 2006 IEEE Computer Society Conference on Computer Vision and
  Pattern Recognition (CVPR'06)}, volume~2, pages 1735--1742. IEEE, 2006.

\bibitem{han2019video}
Tengda Han, Weidi Xie, and Andrew Zisserman.
\newblock Video representation learning by dense predictive coding.
\newblock In {\em Proceedings of the IEEE International Conference on Computer
  Vision Workshops}, pages 0--0, 2019.

\bibitem{haresh2020towards}
Sanjay Haresh, Sateesh Kumar, M~Zeeshan Zia, and Quoc-Huy Tran.
\newblock Towards anomaly detection in dashcam videos.
\newblock In {\em 2020 IEEE Intelligent Vehicles Symposium (IV)}, pages
  1407--1414. IEEE.

\bibitem{he2016deep}
Kaiming He, Xiangyu Zhang, Shaoqing Ren, and Jian Sun.
\newblock Deep residual learning for image recognition.
\newblock In {\em Proceedings of the IEEE conference on computer vision and
  pattern recognition}, pages 770--778, 2016.

\bibitem{hinton1994autoencoders}
Geoffrey~E Hinton and Richard~S Zemel.
\newblock Autoencoders, minimum description length and helmholtz free energy.
\newblock In {\em Advances in neural information processing systems}, pages
  3--10, 1994.

\bibitem{kendall1938new}
Maurice~G Kendall.
\newblock A new measure of rank correlation.
\newblock {\em Biometrika}, 30(1/2):81--93, 1938.

\bibitem{kim2019self}
Dahun Kim, Donghyeon Cho, and In~So Kweon.
\newblock Self-supervised video representation learning with space-time cubic
  puzzles.
\newblock In {\em Proceedings of the AAAI Conference on Artificial
  Intelligence}, volume~33, pages 8545--8552, 2019.

\bibitem{kim2018learning}
Dahun Kim, Donghyeon Cho, Donggeun Yoo, and In~So Kweon.
\newblock Learning image representations by completing damaged jigsaw puzzles.
\newblock In {\em 2018 IEEE Winter Conference on Applications of Computer
  Vision (WACV)}, pages 793--802. IEEE, 2018.

\bibitem{kingma2014adam}
Diederik~P Kingma and Jimmy Ba.
\newblock Adam: A method for stochastic optimization.
\newblock {\em arXiv preprint arXiv:1412.6980}, 2014.

\bibitem{larsson2016learning}
Gustav Larsson, Michael Maire, and Gregory Shakhnarovich.
\newblock Learning representations for automatic colorization.
\newblock In {\em European conference on computer vision}, pages 577--593.
  Springer, 2016.

\bibitem{larsson2017colorization}
Gustav Larsson, Michael Maire, and Gregory Shakhnarovich.
\newblock Colorization as a proxy task for visual understanding.
\newblock In {\em Proceedings of the IEEE Conference on Computer Vision and
  Pattern Recognition}, pages 6874--6883, 2017.

\bibitem{lee2017unsupervised}
Hsin-Ying Lee, Jia-Bin Huang, Maneesh Singh, and Ming-Hsuan Yang.
\newblock Unsupervised representation learning by sorting sequences.
\newblock In {\em Proceedings of the IEEE International Conference on Computer
  Vision}, pages 667--676, 2017.

\bibitem{Li2020MSTCNMT}
S. Li, Yazan~Abu Farha, Yun Liu, Ming-Ming Cheng, and Juergen Gall.
\newblock Ms-tcn++: Multi-stage temporal convolutional network for action
  segmentation.
\newblock {\em IEEE transactions on pattern analysis and machine intelligence},
  PP, 2020.

\bibitem{liu2018leveraging}
Xialei Liu, Joost Van De~Weijer, and Andrew~D Bagdanov.
\newblock Leveraging unlabeled data for crowd counting by learning to rank.
\newblock In {\em Proceedings of the IEEE Conference on Computer Vision and
  Pattern Recognition}, pages 7661--7669, 2018.

\bibitem{maaten2008visualizing}
Laurens van~der Maaten and Geoffrey Hinton.
\newblock Visualizing data using t-sne.
\newblock {\em Journal of machine learning research}, 9(Nov):2579--2605, 2008.

\bibitem{misra2016shuffle}
Ishan Misra, C~Lawrence Zitnick, and Martial Hebert.
\newblock Shuffle and learn: unsupervised learning using temporal order
  verification.
\newblock In {\em European Conference on Computer Vision}, pages 527--544.
  Springer, 2016.

\bibitem{mobahi2009deep}
Hossein Mobahi, Ronan Collobert, and Jason Weston.
\newblock Deep learning from temporal coherence in video.
\newblock In {\em Proceedings of the 26th Annual International Conference on
  Machine Learning}, pages 737--744, 2009.

\bibitem{noroozi2017representation}
Mehdi Noroozi, Hamed Pirsiavash, and Paolo Favaro.
\newblock Representation learning by learning to count.
\newblock In {\em Proceedings of the IEEE International Conference on Computer
  Vision}, pages 5898--5906, 2017.

\bibitem{paszke2017automatic}
Adam Paszke, Sam Gross, Soumith Chintala, Gregory Chanan, Edward Yang, Zachary
  DeVito, Zeming Lin, Alban Desmaison, Luca Antiga, and Adam Lerer.
\newblock Automatic differentiation in pytorch.
\newblock 2017.

\bibitem{purushwalkam2020aligning}
Senthil Purushwalkam, Tian Ye, Saurabh Gupta, and Abhinav Gupta.
\newblock Aligning videos in space and time.
\newblock In {\em European Conference on Computer Vision}, pages 262--278.
  Springer, 2020.

\bibitem{richard2019temporal}
Alexander Richard.
\newblock {\em Temporal Segmentation of Human Actions in Videos}.
\newblock PhD thesis, Universit{\"a}ts-und Landesbibliothek Bonn, 2019.

\bibitem{sermanet2018time}
Pierre Sermanet, Corey Lynch, Yevgen Chebotar, Jasmine Hsu, Eric Jang, Stefan
  Schaal, Sergey Levine, and Google Brain.
\newblock Time-contrastive networks: Self-supervised learning from video.
\newblock In {\em 2018 IEEE International Conference on Robotics and Automation
  (ICRA)}, pages 1134--1141. IEEE, 2018.

\bibitem{srivastava2015unsupervised}
Nitish Srivastava, Elman Mansimov, and Ruslan Salakhudinov.
\newblock Unsupervised learning of video representations using lstms.
\newblock In {\em International conference on machine learning}, pages
  843--852, 2015.

\bibitem{su2017order}
Bing Su and Gang Hua.
\newblock Order-preserving wasserstein distance for sequence matching.
\newblock In {\em Proceedings of the IEEE Conference on Computer Vision and
  Pattern Recognition}, pages 1049--1057, 2017.

\bibitem{sultani2018real}
Waqas Sultani, Chen Chen, and Mubarak Shah.
\newblock Real-world anomaly detection in surveillance videos.
\newblock In {\em Proceedings of the IEEE conference on computer vision and
  pattern recognition}, pages 6479--6488, 2018.

\bibitem{tran2015learning}
Du Tran, Lubomir Bourdev, Rob Fergus, Lorenzo Torresani, and Manohar Paluri.
\newblock Learning spatiotemporal features with 3d convolutional networks.
\newblock In {\em Proceedings of the IEEE international conference on computer
  vision}, pages 4489--4497, 2015.

\bibitem{tran2018closer}
Du Tran, Heng Wang, Lorenzo Torresani, Jamie Ray, Yann LeCun, and Manohar
  Paluri.
\newblock A closer look at spatiotemporal convolutions for action recognition.
\newblock In {\em Proceedings of the IEEE conference on Computer Vision and
  Pattern Recognition}, pages 6450--6459, 2018.

\bibitem{vincent2008extracting}
Pascal Vincent, Hugo Larochelle, Yoshua Bengio, and Pierre-Antoine Manzagol.
\newblock Extracting and composing robust features with denoising autoencoders.
\newblock In {\em Proceedings of the 25th international conference on Machine
  learning}, pages 1096--1103, 2008.

\bibitem{vondrick2016generating}
Carl Vondrick, Hamed Pirsiavash, and Antonio Torralba.
\newblock Generating videos with scene dynamics.
\newblock In {\em Advances in neural information processing systems}, pages
  613--621, 2016.

\bibitem{wang2018non}
Xiaolong Wang, Ross Girshick, Abhinav Gupta, and Kaiming He.
\newblock Non-local neural networks.
\newblock In {\em Proceedings of the IEEE conference on computer vision and
  pattern recognition}, pages 7794--7803, 2018.

\bibitem{wei2018learning}
Donglai Wei, Joseph~J Lim, Andrew Zisserman, and William~T Freeman.
\newblock Learning and using the arrow of time.
\newblock In {\em Proceedings of the IEEE Conference on Computer Vision and
  Pattern Recognition}, pages 8052--8060, 2018.

\bibitem{wiki:Kendall_rank_correlation_coefficient}
Wikipedia.
\newblock {Kendall rank correlation coefficient} --- {W}ikipedia{,} the free
  encyclopedia, 2020.

\bibitem{wiki:Coefficient_of_determination}
{Wikipedia contributors}.
\newblock Coefficient of determination --- {Wikipedia}{,} the free
  encyclopedia, 2021.
\newblock [Online; accessed 22-March-2021].

\bibitem{xu2019self}
Dejing Xu, Jun Xiao, Zhou Zhao, Jian Shao, Di Xie, and Yueting Zhuang.
\newblock Self-supervised spatiotemporal learning via video clip order
  prediction.
\newblock In {\em Proceedings of the IEEE Conference on Computer Vision and
  Pattern Recognition}, pages 10334--10343, 2019.

\bibitem{zhang2013actemes}
Weiyu Zhang, Menglong Zhu, and Konstantinos~G Derpanis.
\newblock From actemes to action: A strongly-supervised representation for
  detailed action understanding.
\newblock In {\em Proceedings of the IEEE International Conference on Computer
  Vision}, pages 2248--2255, 2013.

\bibitem{zou2012deep}
Will Zou, Shenghuo Zhu, Kai Yu, and Andrew~Y Ng.
\newblock Deep learning of invariant features via simulated fixations in video.
\newblock In {\em Advances in neural information processing systems}, pages
  3203--3211, 2012.

\bibitem{zou2011unsupervised}
Will~Y Zou, Andrew~Y Ng, and Kai Yu.
\newblock Unsupervised learning of visual invariance with temporal coherence.
\newblock In {\em NIPS 2011 workshop on deep learning and unsupervised feature
  learning}, volume~3, 2011.

\end{thebibliography}
}

\end{document}